\documentclass[letterpaper, 10 pt, journal, twoside]{IEEEtran}
\IEEEoverridecommandlockouts                              %

\usepackage{times}
\usepackage[dvipsnames]{xcolor}

\usepackage[
    maxnames=1,
    minnames=1,
    style=ieee]{biblatex}
\AtEveryBibitem{%
  \clearlist{publisher} %
  \clearlist{organization} %
  \clearfield{issn} %
  \clearfield{doi} %
  \clearfield{isbn} %
  \clearfield{pages} %
  \clearfield{volume} %
  \clearfield{month} %
  \clearfield{number} %

  \ifentrytype{online}{}{%
    \clearfield{url}
  }
}

\DeclareSourcemap{
    \maps[datatype=bibtex]{
      \map{
           \step[fieldsource=journal, match=Robotics: Science and Systems, replace=\regexp{RSS}]
           \step[fieldsource=booktitle, match=Robotics: Science and Systems, replace=\regexp{RSS}]
           \step[fieldsource=journal, match=International Conference on Robotics and Automation, replace=\regexp{ICRA}]
           \step[fieldsource=booktitle, match=International Conference on Robotics and Automation, replace=\regexp{ICRA}]
           \step[fieldsource=journal, match=International Conference on Learning Representations, replace=\regexp{ICLR}]
           \step[fieldsource=booktitle, match=International Conference on Learning Representations, replace=\regexp{ICLR}]
           \step[fieldsource=booktitle, match=European Conference on Computer Vision, replace=\regexp{ECCV}]
           \step[fieldsource=journal, match=European Conference on Computer Vision, replace=\regexp{ECCV}]
           \step[fieldsource=booktitle, match=International Conference on Computer Vision, replace=\regexp{ICCV}]
           \step[fieldsource=journal, match=International Conference on Computer Vision, replace=\regexp{ICCV}]
           \step[fieldsource=booktitle, match=IEEE Transactions on Pattern Analysis and Machine Intelligence, replace=\regexp{PAMI}]
           \step[fieldsource=journal, match=IEEE Transactions on Pattern Analysis and Machine Intelligence, replace=\regexp{PAMI}]
           \step[fieldsource=booktitle, match=International Conference on Intelligent Robots and Systems, replace=\regexp{IROS}]
           \step[fieldsource=booktitle, match=Proceedings of the IEEE/CVF Conference on Computer Vision and Pattern Recognition, replace=\regexp{CVPR}]
           \step[fieldsource=booktitle, match=Conference on Computer Vision and Pattern Recognition, replace=\regexp{CVPR}]
           \step[fieldsource=journal, match=\regexp{International}, replace=\regexp{Int.}]
           \step[fieldsource=booktitle, match=\regexp{International}, replace=\regexp{Int.}]
           \step[fieldsource=journal, match=\regexp{Conference}, replace=\regexp{Conf.}]
           \step[fieldsource=booktitle, match=\regexp{Conference}, replace=\regexp{Conf.}]
           \step[fieldsource=journal, match=\regexp{Transactions}, replace=\regexp{Trans.}]
           \step[fieldsource=booktitle, match=\regexp{Transactions}, replace=\regexp{Trans.}]
           \step[fieldsource=booktitle, match=Robotics and Automation Letters, replace=\regexp{RA-L}]
           \step[fieldsource=journal, match=Robotics and Automation Letters, replace=\regexp{RA-L}]
           \step[fieldsource=booktitle, match=Advances in Neural Information Processing Systems, replace=\regexp{NeurIPS}]
           \step[fieldsource=journal, match=Advances in Neural Information Processing Systems, replace=\regexp{NeurIPS}]
           \step[fieldsource=booktitle, match=Neural Information Processing Systems, replace=\regexp{NeurIPS}]
           \step[fieldsource=journal, match=Neural Information Processing Systems, replace=\regexp{NeurIPS}]
          }
    }     
}

\AtBeginBibliography{\scriptsize}
\usepackage{multicol}
\usepackage[hidelinks]{hyperref}
\hypersetup{
    colorlinks=true,
    linkcolor=black,
    filecolor=black,      
    urlcolor=orange,
    citecolor=black,
}
\let\labelindent\relax
\usepackage{enumitem}
\usepackage{lipsum}
\usepackage{graphicx}
\usepackage{amsfonts}
\usepackage{booktabs}
\usepackage{pifont}
\usepackage{amsmath}
\usepackage{siunitx}
\usepackage{multirow}
\usepackage{comment}
\usepackage{stfloats}
\usepackage{subcaption}
\usepackage[font=footnotesize]{caption} 
\usepackage[capitalise, noabbrev]{cleveref}

\def\HideCuts{0} %
\def\HideTodo{0} %

\usepackage{xspace}
\makeatletter
\DeclareRobustCommand\onedot{\futurelet\@let@token\@onedot}
\def\@onedot{\ifx\@let@token.\else.\null\fi\xspace}
\def\etal{\emph{et al}\onedot}
\makeatother

\newcommand{\todo}[1]{\if\HideTodo0{\color{red} [todo: #1]}\fi}
\newcommand{\cut}[1]{\if\HideCuts0{\color{gray} [cut: #1]}\fi}
\newcommand{\rom}[1]{\uppercase\expandafter{\romannumeral #1\relax}}

\newcommand{\cmark}{\ding{51}}
\newcommand{\xmark}{\ding{55}}

\newcommand{\hstlong}{Human Scene Transformer}
\newcommand{\hst}{HST}
\newcommand{\humanpose}{human pose} %

\renewcommand{\baselinestretch}{0.95}

\begin{document}

\title{Robots That Can See:\\Leveraging Human Pose for Trajectory Prediction}

\author{Tim Salzmann$^{1, 2}$, Hao-Tien Lewis Chiang$^{1}$, Markus Ryll$^{2}$, Dorsa Sadigh$^{1, 3}$, Carolina Parada$^{1}$, and Alex Bewley$^{1}$%
\thanks{Manuscript received: May 2nd, 2023; Revised: July 21st, 2023; Accepted: August 21st, 2023.}%
\thanks{This paper was recommended for publication by Gentiane Venture upon evaluation of the Associate Editor and Reviewers’ comments.}%
\thanks{$^{1}$Tim Salzmann, Hao-Tien Lewis Chiang, Dorsa Sadigh, Carolina Parada, and Alex Bewley are with Google DeepMind {\tt\small \{tsal, lewispro, dorsas, carolinap, bewley\}@google.com}.}%
\thanks{$^{2}$Tim Salzmann and Markus Ryll are with Technical University Munich {\tt\small \{tim.salzmann, markus.ryll\}@tum.de}.}%
\thanks{$^{3}$Dorsa Sadigh is with Stanford University {\tt\small dorsa@cs.stanford.edu}}%
\thanks{Digital Object Identifier (DOI): see top of this page.}%
}

\markboth{IEEE Robotics and Automation Letters. Preprint Version. Accepted August, 2023}
{Salzmann \MakeLowercase{\textit{et al.}}: Robots That Can See}

\maketitle

\begin{abstract}
Anticipating the motion of all humans in dynamic environments such as homes and offices is critical to enable safe and effective robot navigation.
Such spaces remain challenging as humans do not follow strict rules of motion and there are often multiple occluded entry points such as corners and doors that create opportunities for sudden encounters. In this work, we present a Transformer based architecture to predict human future trajectories in human-centric environments from input features including human positions, head orientations, and 3D skeletal keypoints from onboard in-the-wild sensory information. The resulting model captures the inherent uncertainty for future human trajectory prediction and achieves state-of-the-art performance on common prediction benchmarks and a human tracking dataset captured from a mobile robot adapted for the prediction task. Furthermore, we identify new agents with limited historical data as a major contributor to error and demonstrate the complementary nature of 3D skeletal poses in reducing prediction error in such challenging scenarios.

\vspace{0.3em}
\noindent Project page: \url{https://human-scene-transformer.github.io/}

\end{abstract}


\IEEEpeerreviewmaketitle

\section{Introduction}

\IEEEPARstart{T}{he} presence of robots sharing the environment with humans has led to a need for effective methods to understand the human's intention in order to avoid collisions and ensure smooth interactions between humans and robots. A series of steps are required for a robot to successfully navigate around humans in dynamic scenes, namely perception, prediction and planning. Perception is responsible for detecting the presence of humans and extracting other features of the environment around the robot. Prediction aims to model how humans will move into the future, and finally planning future actions towards a goal while avoiding collisions. %
In this work we focus on the prediction step where the inputs include perceived visual features and the outputs are predicted trajectory distributions for motion planning.

While the trajectory prediction topic has been extensively studied in the context of autonomous driving~\cite{yuan2021agentformer, nayakanti2022wayformer, ngiam2021scene, salzmann2020trajectron++}, predicting human trajectories in other environments where service robots could have a profound impact, such as offices, homes, hospitals, and elderly care facilities, have received less attention. 
In contrast to street scenes, the main dynamic agent in service robotics environments are humans carrying out a wider range of tasks that require diverse motions in unstructured environments, compared to the setting of driving which has significant structure in place such as staying within lanes or following the traffic rules. Additionally, the spatial environment is generally smaller with a higher degree of perceptual obstructions (e.g. blind corners or internal walls) resulting in a closer proximity upon first observation due to occluded entry-points.
Our goal is to enable more natural, safe, smooth, and predictable navigation by anticipating where humans will be moving in the near future using the robot’s onboard sensors.
\begin{figure}
    \centering
    \includegraphics[width=\linewidth,trim={0 1cm 0 1.5cm},clip]{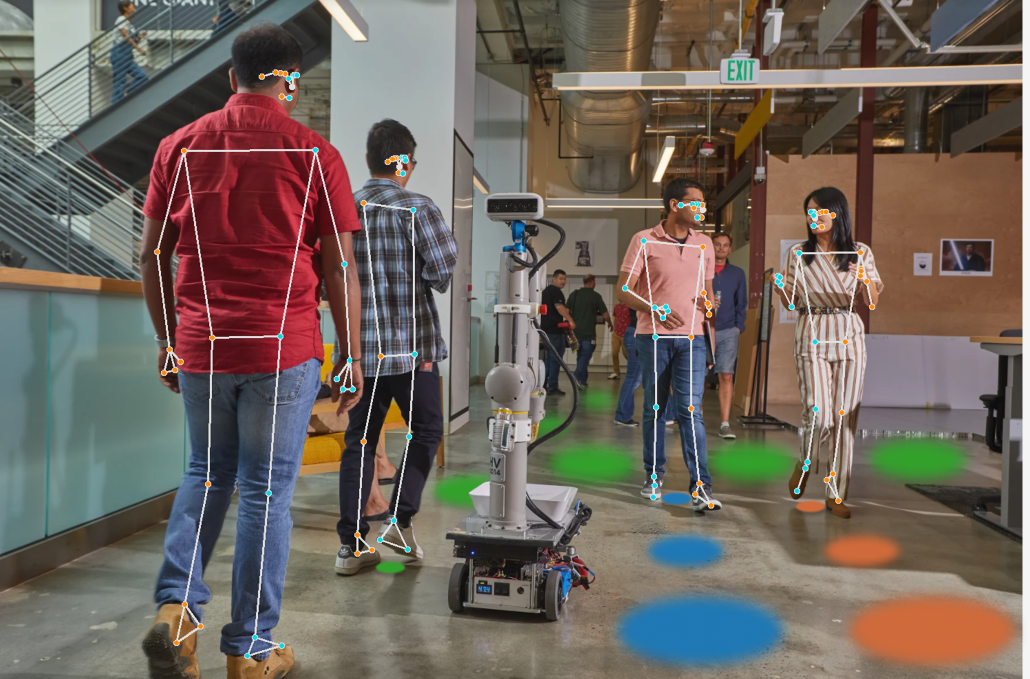}
    \caption{ A service robot navigating a busy office space. To do so it anticipates human motion using human-position and visual features such as head orientation or 3D skeletal keypoints.\vspace{1.5em}} %
    \label{fig:hero}
\end{figure}
Several existing methods %
try to generalize vehicle trajectory prediction ideas to human trajectory prediction by representing humans as 2D bounding boxes~\cite{pellegrini2009you, lerner2007crowds, mangalam2020not, sadeghian2019sophie, mangalam2021goals}.
While bounding boxes can be sufficient for predicting vehicle trajectories in outdoor environments~\cite{ngiam2021scene, ivanovic2019trajectron, salzmann2020trajectron++, yuan2021agentformer}, the reduction of human agents to bounding boxes neglects the plethora of perceptual information, present in a human-centric scene -- scenes with one or multiple humans sharing limited space with the robot --- such as a robot navigating in a hallway of a busy office as in Fig.~\ref{fig:hero}. 
In such settings, humans tend to take advantage of information beyond each other's position: There is substantial information about people's intent when \emph{they turn their head} or \emph{look at where to walk next} --- humans predict and anticipate each other's intentions through vision-based features: pose, gaze, and gestures. These visual cues are also apparent for a human viewing a static image absent of temporal cues.
Similarly, we will demonstrate that a robot navigating in close proximity to humans as in Fig.~\ref{fig:hero} 
enables a more detailed model of the human beyond simple bounding boxes and leads to more accurate prediction of human trajectories. Specifically, we posit the research question: \emph{``Can information from human visual features lead to improved prediction accuracy?''}

\newpage
We present the \hstlong\ (\hst) which leverages different feature streams: Historic positions of each human, vision-based features such as skeletal keypoints (see \cref{fig:hero}, joints of the human skeleton) or head orientation when available. %
We specifically focus on demonstrating the usefulness of noisy in-the-wild human skeletal information from a 3D human pose estimator.

\noindent As such our contribution is threefold:
\begin{itemize}[leftmargin=*]
    \item To the best of our knowledge, we are the first to adapt the trajectory prediction task to the domain of human-centric navigation and demonstrate that 3D vision-based features improve prediction performance in a service robot context notwithstanding imperfect in-the-wild data (without a dedicated motion capture system). Specifically in the regime of limited historical data, which is particularly under-explored, but a common scenario in indoor robot navigation our key idea is to use 3D vision features as complimentary predictive cues.
    \item We present a prediction architecture, which flexibly processes and includes detailed vision-based human features such as skeletal keypoints and head orientation. To target crowded human-centric environments, we define HST within a system of components from fields of Computer Vision, Machine Learning, and Autonomous Driving to make use of real-world sensor data instead of relying on ground truth labels. We demonstrate \hst's capability to consistently model interactions which is critical in human-centric environments.
    \item We highlight a gap in prior work by showing the limitations of existing datasets for human trajectory prediction in indoor navigation. We propose an adaptation of existing datasets that can enable this new way of future trajectory prediction. Using this adaptation, we demonstrate the feasibility of our approach in human-centric environments. Simultaneously, we display state-of-the art performance on a common outdoor pedestrian prediction dataset.
\end{itemize}

\section{Related Work}
Predicting the future trajectory of humans is a challenging task where prior work has considered various motion models, scene context, and social interaction. We will revisit three research fields which influence our targeted domain of service robots and the use of vision-based human skeletal (human pose) features. We will first outline current research in \emph{Trajectory Prediction}, where the center position of an agent (which can be a human but also a human driven vehicle) is forecasted over time. Subsequently, we will introduce work in the field of \emph{Human Pose Prediction}, which uses a skeletal representation of the human and predicts this pose over time. Finally, we outline approaches in \emph{Pose Estimation}, where the estimated human pose is directly used in trajectory prediction in application domains outside of human-centric navigation.

\smallskip \noindent \textbf{Trajectory Prediction.} From the early work of Pellegrini \etal \cite{pellegrini2009you} for short-term future locations of humans for next frame data association to the recent longer term multi-second prediction methods that are often used in autonomous driving \cite{salzmann2020trajectron++, ngiam2021scene, nayakanti2022wayformer, yuan2021agentformer, czech2022board}, trajectory prediction research has played an important role in improving down-stream robotic tasks. 
Prior approaches consider scene context \cite{salzmann2020trajectron++}, motion dynamics \cite{salzmann2020trajectron++, ngiam2021scene}, and the interaction between agents \cite{ngiam2021scene}. Salzmann \etal~\cite{salzmann2020trajectron++} combine historic agent positions with scene and dynamics constraints to make informed predictions. In an autonomous driving context, extracting additional visual information about the human actor (driver, cyclist, pedestrian) is challenging due to occlusion (driver only partially visible in the car) or distance. Therefore, the information representing each agent is reduced to a position estimate per observation timestep~\cite{salzmann2020trajectron++, ngiam2021scene, nayakanti2022wayformer, yuan2021agentformer}.  Unlike the large prediction range required for self-driving~\cite{sun2021rsn}, we focus on service-robot environments where people are generally close enough to the robot to obtain a richer visual representation of the human. As such, our work can benefit from recent approaches that fuse LiDAR information adapting the \humanpose\ estimation problem to a robotic sensor suit \cite{zanfir2022hum3dil}.
While prior works in trajectory prediction rely on Generative Adversarial Networks (GANs)~\cite{gupta2018social, sadeghian2019sophie} or Conditional Variational Autoencoders (CVAEs)~\cite{mangalam2020not, ivanovic2020multimodal, salzmann2020trajectron++, ivanovic2019trajectron, ivanovic2018generative}, this work follows the recent trend towards Transformer architectures as they naturally lend themselves to the set-to-set prediction problems such as multi-agent trajectory prediction and are invariant to a varying number of agents. 
Specifically, we leverage the fundamental idea of a Transformer based prediction framework~\cite{ngiam2021scene, yuan2021agentformer, nayakanti2022wayformer} inspired by Ngiam \etal~\cite{ngiam2021scene}. Their Transformer architecture is used for vehicle trajectory prediction in autonomous driving applications and captures joint interactions between vehicles.

\smallskip \noindent \textbf{Pose Prediction.} 
Another related area is \humanpose\ forecasting in 3D~\cite{corona2020context, yuan2020dlow, zhang2021we, mao2020history, salzmann2022motron}. Corona \etal use scene context in the form of an influence graph to refine a future trajectory of 3D \humanpose s for a single subject in a motion capture environment \cite{corona2020context}. For multi-person pose prediction, 
\cite{vendrow2022somoformer} %
extends DeTR \cite{carion2020end} to predict multi-person 2D poses from a single image.
Towards social robot navigation, Narayanan \etal use a sequence of \humanpose s, also known as gait, to classify a person's emotional state for setting appropriate proximal distance constraints \cite{narayanan2020proxemo}. However, these approaches commonly consider a single human motion relying on ground truth pose information from a motion capture system, while we target multi-human in-the-wild scenarios which are not limited to spaces with a motion capture system.

\smallskip \noindent \textbf{Pose Estimation in Trajectory Prediction.} 
There have been prior efforts to combine pose estimation with trajectory prediction, i.e., informing forecasted trajectories by incorporating historic pose information. However, these works are either limited to prediction in 2D image space~\cite{yagi2018future, chen2020pedestrian, czech2022board} or operating on motion capture datasets which do not exhibit diverse positional movement of the human~\cite{corona2020context, mahdavian2022stpotr}. 
Yagi \etal \cite{yagi2018future} showed that augmenting a convolutional auto-encoder style model with scale and pose encoders reduces prediction error compared to position only; however, their approach is applied to first-person video using 2D pose detection and limiting the prediction to the 2D image space.
Similarly, Chen \etal \cite{chen2020pedestrian} use a convolutional and recurrent architecture to segment an image into heterogeneous traffic objects and body parts, before using a Transformer decoder to attend to feature maps and extract objects leading to improvements in 2D image space prediction. When considering predicting multiple poses, Adeli \etal \cite{adeli2021tripod} use a form of graph attention to capture dependencies between interacting agents but this work is, again, limited to 2D first-person videos. Other works have explored attention mechanisms between multiple human features in the image-view~\cite{czech2022board}.
However, for robotic navigation it is desired to obtain predictions for agents across multiple sensors and ideally in a 3D or bird's-eye metric space. In this work, we follow these requirements by solely relying on onboard sensor information of a robotic platform and predict in the metric global frame rather than in image space.

\section{\hstlong}\label{sec:model}

Our proposed method \hstlong\ (\hst) follows the concept of masked sequence to sequence prediction using an architecture with Transformer blocks (see \cref{fig:model_diagram} - top right). This approach has shown promising vehicle prediction results in the autonomous driving domain \cite{ngiam2021scene}. \hst\ introduces multiple important ideas extending the general Transformer architecture which makes it suitable for trajectory prediction in human-centric environments. These include the utilization of vision-based human features (\cref{sec:input}), a feature attention mechanism to merge multiple potentially incomplete features (\cref{sec:arch} - Input Embedding), an improved attention mechanism facilitating a more complete information flow (\cref{sec:arch} - Full Self-Attention), and a self-alignment layer which elegantly solves the problem of discriminating between multiple masked agent timesteps while keeping permutation equivariance (\cref{sec:arch} - Agent Self-Alignment). Notably, implementing an attention based architecture end-to-end, the model is agnostic to the number of humans per frame. This means the model can dynamically handle a varying number of humans in different timesteps during inference. The maximum number of jointly (single forward pass) predicted humans is only limited by available memory.

\subsection{Model Inputs: Incorporating Vision-based Features}\label{sec:input}
\begin{figure*}[t]
    \centering
    \begin{minipage}[t]{0.68\textwidth}
        \vspace{-0em}
        \includegraphics[width=\textwidth]{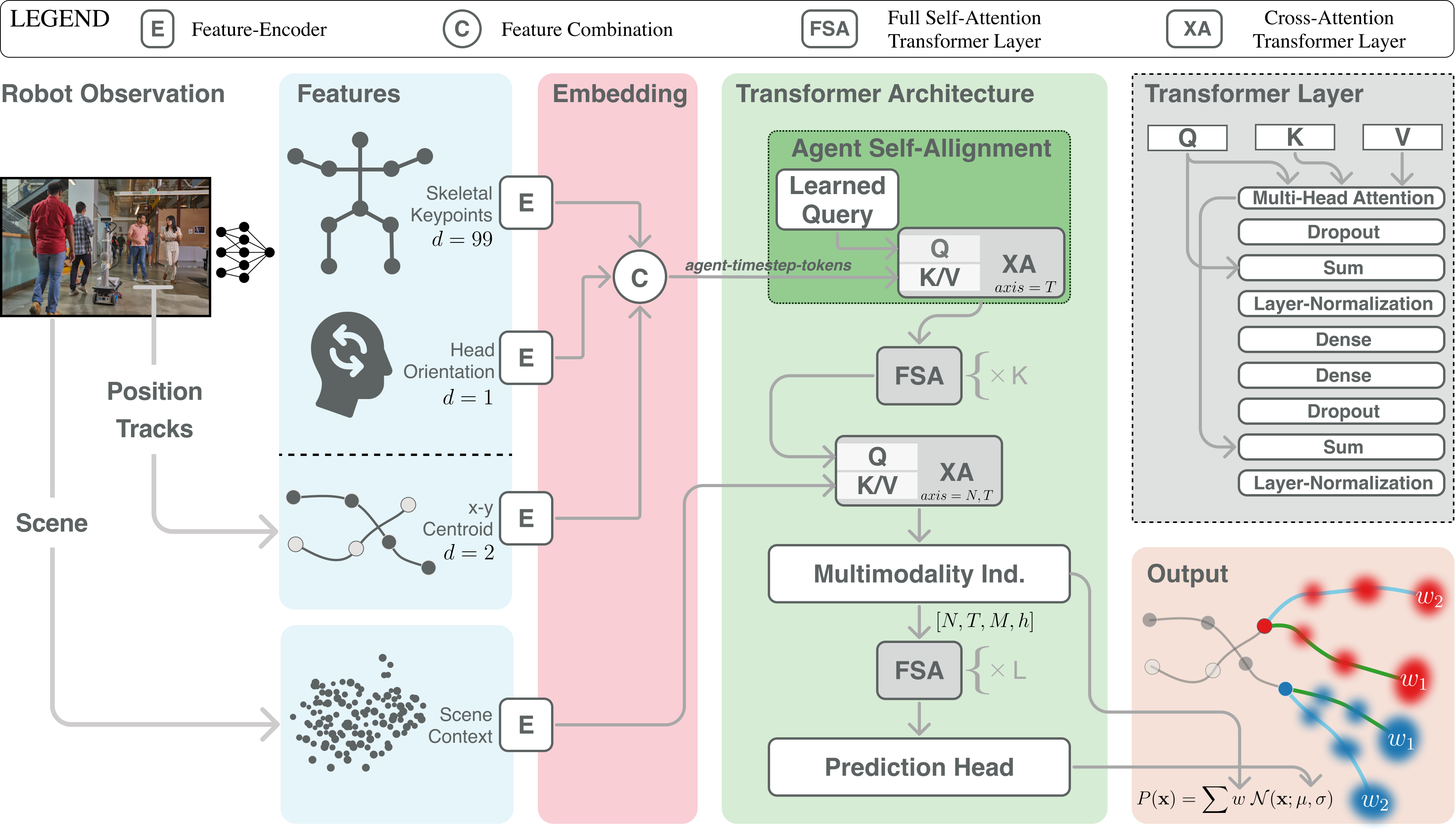}
    \end{minipage}\hfill
    \begin{minipage}[t]{0.31\textwidth}
        \caption{\textbf{Overview of the \hst\ architecture.} From the robot's sensors we extract the scene context, the historic position tracks of each agent, and vision based skeletal keypoints/head orientation when feasible. All features are encoded individually before the agent features are combined via cross-attention (XA) using a learned query tensor. The resulting agent-timestep-tokens is passed to our Agent Self-Alignment layer which enables the use of subsequent full self-attention (FSA) layers. Embedded scene context is attended to via cross-attention (XA). After multimodality is induced and further FSA layers the model outputs the parameters of a Normal distribution for each agent at each prediction timestep. We can represent the full output structure as a Gaussian Mixture Model (formula in bottom right) over all possible futures where the mixture coefficients $w$ come from the Multimodality Induction. Both cross-attention (XA) and full self-attention layers use the Transformer layer (top right) with different input configurations for Query (Q), Key (K), and Value (V).}\label{fig:model_diagram}
    \end{minipage}
\end{figure*}

The robot's observations for the last $H+1$ timesteps are processed as agent features and scene context (\cref{fig:model_diagram} blue box).
The scene context can be an occupancy grid or a raw point cloud at the current timestep, containing information common to all nearby agents (e.g. static obstacles).
Agent features include the centroid position and vision-based features: skeletal keypoints, and head orientation for each agent.

\begin{figure}
    \centering
    \includegraphics[width=\linewidth,trim={0 6cm 6.0cm 0cm},clip]{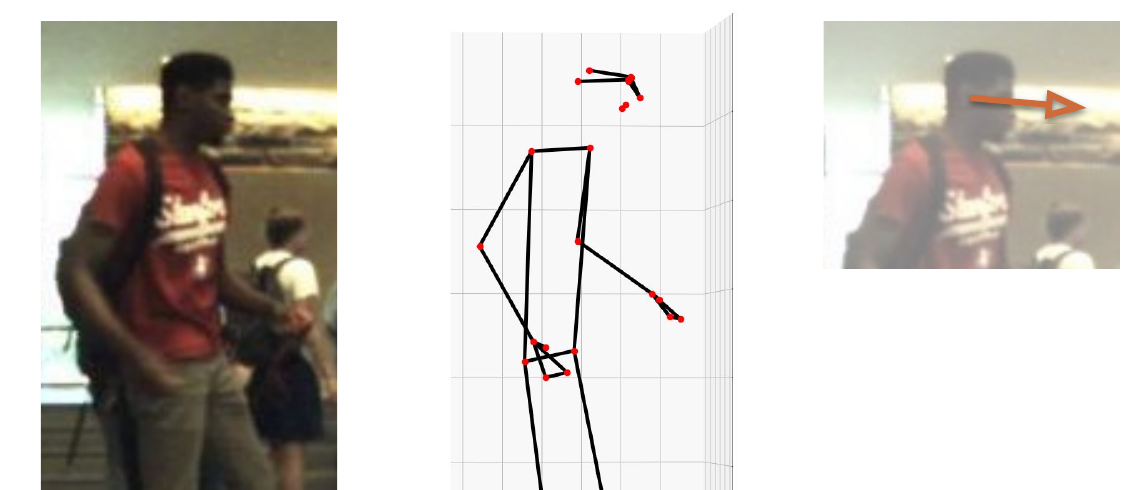}
    \caption{\textbf{Process of estimating three dimensional vision-based features of the human.} (Left) Image of the human cropped based on the bounding box detection. Lighting conditions can be sub-optimal in an human-centric environment. (Middle) Inferred three dimensional pose from the trained pose estimator. (Right) Head orientation post-processed from the pose keypoints.}
    \label{fig:data_process}
\end{figure}

To extract these vision-based features from the raw data, image patches for all agents are first obtained by projecting their detected 3D bounding boxes into the 360 degree image using ex- and intrinsic camera calibrations (see \cref{fig:data_process}-a).
To extract skeletal key points from these patches, one could choose from a plethora of off-the-shelf skeletal keypoints extractor for images~\cite{8765346, fang2022alphapose, maji2022yolo, movenet, posenet}.
However, these extractors commonly output keypoints in a 2D image coordinate frame. 
To produce 3D keypoints, we follow the work of Grishchenko \etal~\cite{grishchenko2022blazepose} to estimate 3D keypoints from images using a pre-trained model: 
As existing datasets commonly only include 2D keypoint annotations, the 3D label required for supervised pre-training is generated by fitting a parametric human shape model to available 2D keypoints solving the following optimization problem:
\begin{equation}\label{eq:opt}
    \small
     \text{argmin}_{\mathbf{k}} \left(\lVert r(\mathbf{k}) - \hat{\mathbf{k}}_2 \rVert_2 + \lambda p(\mathbf{k})\right) ,
\end{equation}
where $\mathbf{k}$ are the 3D skeletal keypoints, $\hat{\mathbf{k}}_2$ is the 2D keypoints label, $r: \mathbb{R}^{33 \times 3} \rightarrow \mathbb{R}^{33 \times 2}$ is the re-projection function projecting 3D points into the 2D image space using the camera calibrations. Many representations (from $<20$~\cite{lin2014microsoft} to $>500$~\cite{mp_holistic} keypoints) are present in the literature; we settled for a 33 keypoints skeleton representation~\cite{grishchenko2022blazepose, xu2020ghum} as it presents itself as a minimal representation capturing \emph{both} head information and limb articulation. The learned prior distribution over human pose configuration $p(\mathbf{k})$ penalizes infeasible poses which can arise in optimization for the underdetermined 3D-2D-projection problem, as multiple 3D poses can result in equivalent 2D projections. A infeasible pose would be a configuration of human joints which is physically infeasible for a human to achieve (e.g. head rotated full 180$^{\circ}$). The prior distribution is learned by fitting a variational autoencoder to a dataset of feasible 3D human poses.
The 3D keypoints in the camera space are transformed to the global coordinate frame using extrinsic camera calibration.
Given the skeletal keypoints $\mathbf{k}$ from optimizing \cref{eq:opt}, we can easily extract the head orientation as the vector from in-between ears to in-between eyes. 
Note that vision-based agent features may not be available for all agents at all timesteps. 
This can be due to the agent being too far away to reliably extract keypoints (e.g., the picture being too small). We will show in \cref{sec:arch} - Input Embedding how we can leverage inherent properties of our \hst\ Transformer architecture to deal with these situations.

\subsection{Model Architecture}\label{sec:arch}
\cref{fig:model_diagram} outlines the \hst\ architecture. We will explain the individual components in this section by first introducing the Transformer layer as core concept and subsequently following the data flow depicted in \cref{fig:model_diagram}.

\smallskip \noindent \textbf{Transformer Layer.}
The primary building block of the model's architecture is the Transformer layer (shown in \cref{fig:model_diagram} top right), which itself is comprised of a Multi-Head Attention layer~\cite{vaswani2017attention} and multiple dense and normalization layers. The Transformer layer receives three tensors as input: Query (Q), Key (K), and Value (V). However, a single tensor may be used for multiple of the inputs. Consequently, we define a self-attention (SA) operation as a Transformer Layer where inputs Q, K, and V are the same tensor: The tensor attends to itself and conveys its information along one or more dimensions. Similarly, we define cross-attention (XA) as a Transformer Layer where the Q input is distinct from the K/V inputs. Intuitively the query attends to additional information from a different tensor as means of merging multiple streams of information.
For a comprehensive explanation on the Attention mechanism and its inputs we refer the reader to Vaswani \etal~\cite{vaswani2017attention}.

\begin{figure}[t]
    \centering
    \begin{minipage}[c]{0.55\linewidth}
        \includegraphics[width=\textwidth]{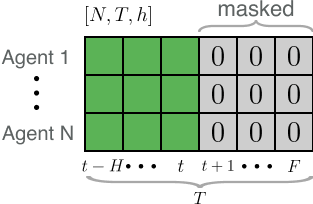}
    \end{minipage}\hfill
    \begin{minipage}[c]{0.40\linewidth}
        \caption{Structure and masking of embedded tensors. Future agent-timestep-tokens are masked and subsequently filled by the Transformer structure, iteratively with updated latent representations and finally with position distribution information on the output level.}\label{fig:tensor}
    \end{minipage}
\end{figure}

\smallskip \noindent \textbf{Input Embedding.}
The input agent features (blue) are tensors of shape $[N, T, d]$, where $d=2$ for the x-y centroid position, $d=99$ for the x-y-z position of 33 skeletal keypoints, and $d=1$ for the head orientation. 
These tensors contain information of all $N$ nearby agents for all $H + 1$ historic and current input timesteps.
If an agent's feature is not observed at specific timesteps, we mask those timesteps with $0$.
As depicted in \cref{fig:tensor}, we also mask all future $F$ timesteps for all agents by setting their feature value to $0$, thus making only historical and current information available to the model.
This masking approach is a well known technique in missing-data problems such as future prediction using Transformer based architectures~\cite{vaswani2017attention, ngiam2021scene, yuan2021agentformer}. Masking exploits the inductive bias inherent in the prediction problem, which allows for the filling of the missing information using available context in the vicinity of the gaps. As such, our approach allows for missing keypoints in frames due to bad lighting or other influences as the Transformer effectively ``fills'' in for the missing information.

The agent features are encoded independently and are combined by a learned attention query. This masked attention mechanism offers scalability to systems that have a large number of features with limited availability.
The combined agent features result in a latent tensor for each agent and timestep of shape $[N, T, h]$ where $T = H + 1 + F$ and $h$ is the size of the token dimension. For simplicity we will refer to such tensors of latent representations throughout the network as \emph{agent-timestep-tokens}.

\smallskip \noindent \textbf{Full Self-Attention Via Agent Self-Alignment.}
In contrast to previous methods~\cite{ngiam2021scene} which alternately attends to agents and time dimensions separately via factorized attention, we propose a \emph{full self-attention} (FSA) operation where each agent-timestep-token attends to all agent-timestep-tokens along \emph{both} agent and time dimensions. This provides a more direct path of information flow. For example in social interactions, a change in action such as adjustment in walking direction does not have an immediate influence on other humans in proximity but rather influences their future. Following this illustration, an agent at a given timestep in our Transformer architecture should be able to attend not just to other agents at the current timestep (factorized attention) but to \emph{all agents} at \emph{all timesteps} (full self-attention).

Na\"ively using full-self-attention can lead to sub-optimal outcomes. Since all future agent-timestep-tokens are masked out, two agents with the same masked future agent-timestep-tokens will also have the same input (Query) representation to the Transformer layer (\cref{fig:model_diagram} top right). %
This prevents the model from associating future timesteps of an agent with the agent's history, since all future agents' timesteps ``look'' the same (masked).
The problem could be addressed by enforcing an innate order on perceived agents, where all agents are enumerated. This, however, would eliminate the permutation invariant set-to-set prediction capabilities~\cite{lee2019set}; one of the core strengths of Transformers: An agent's future would be predicted differently based on its enumeration embedding with the same historic features.

Instead, we solve this problem and achieve full self-attention via a simple approach that we refer to as \emph{agent self-alignment}. 
The agent-timestep-tokens resulting from the feature combination are cross-attended with a learned query tensor \textit{only} in the time dimension. This query, a weight matrix jointly optimized with all other network weights during training, learns to propagate available historic information for each agent to future timesteps, enabling the model to align future masked timesteps of an agent with historic ones during full self-attention without an explicit enumeration embedding. After this process, which is visualized in the dark green box in \cref{fig:model_diagram}, the previously masked future agent-timestep-tokens hold information from the respective agent's history; differentiating them from another.
As such, the agent self-alignment mechanism preserves agents' permutation equivariance and enables full self-attention without restricting information flow along matching timesteps~\cite{ngiam2021scene} or utilizing special attention matrices which explicitly separates agents~\cite{yuan2021agentformer}.
The output agent-timestep-tokens of the agent self-alignment then passes through $K$ transformer layers with full self-attention across agent and time dimensions before cross-attending to the encoded scene features.

\smallskip \noindent \textbf{Multimodality Induction.}
Our architecture can predict multiple consistent futures (modes) for a scene.
To do so, the Multimodality Induction module repeats the agent-timestep-tokens by the number of future modes ($M$), resulting in a tensor of shape $[N, T, M, h]$. To discriminate between modes it is combined with a learned \emph{mode-identifier} tensor of shape $[1, 1, M, h]$. Each future's logit probability $w_m; \; m \in {1, \dots, M}$ is also inferred here by having the \emph{mode-identifier} attend to the repeated input; resulting in shape $[1, 1, M, h]$ which is subsequently reduced by a MLP to output $w_m$ as $[1, 1, M, 1]$.

\smallskip \noindent \textbf{Prediction Head.} The agent-timestep-tokens updated with the learned mode-identifier go through $L$ Transformer layers, again with full self-attention, before predicting per mode parameters $\mu, \sigma$ using a dense layer as \emph{prediction head}.

\subsection{Producing Multimodal Trajectory Distributions}
Combining $\mu$ and $\sigma$ with the mode likelihoods $w_m$ coming from the Multimodality Induction, the \hstlong\ models the distribution of the $i$-th agent's centroid position at each timestep $t$ with a 2D Gaussian Mixture Model (GMM):
\begin{equation}
    \small
    P^i_\theta(\mathbf{x}_t | O(t), ..., O(t-H)) = \sum_{m = 1}^M w_m \mathcal{N}(\mathbf{x}; \mathbf{\sigma}_{m, i, t}, \mathbf{\mu}_{m, i, t}),
    \label{eq:gmm}
\end{equation}
where $m$ is the $m$-th future mode.
Here, we represent the position of an agent at a specific timestep by a GMM with mixture weights $w$ equal to the probability distribution of future modes.

We adopt a joint future loss function, that is, the cumulative negative log-likelihood of the Gaussian mode ($m^*$) with the smallest mean negative log-likelihood:
\begin{equation}
    \small
    \mathcal{L}_\text{minNLL} = \sum_{i, t} -\text{log}(\mathcal{N}(\mathbf{x}^*_{i, t}; \mathbf{\sigma}_{m^*, i, t}, \mathbf{\mu}_{m^*, i, t})),
    \label{eq:minNLLLoss}
\end{equation}
where
\begin{equation}
    \small
    m^* = \text{argmin}_m( \sum_{i, t} -\text{log}(\mathcal{N}(\mathbf{x}^*_{i, t} ; \mathbf{\sigma}_{m, i, t}, \mathbf{\mu}_{m, i, t}))),
    \label{eq:minADEMode}
\end{equation}
and $\mathbf{x}^*_{i, t}$ is the ground truth agent position.
The resulting prediction represents $M$ possible future realizations of all agents at once in a consistent manner, where the mode mixture weights $w$ are shared by all agents in the scene. The most likely future mode during inference is given as
\begin{equation}
    \small
    m^+ = \text{argmax}_m(w_m).
\end{equation}

\section{Experiments}
We structure our experiments to support our contributions: First, we qualitatively and quantitatively demonstrate how our architecture provides accurate predictions for the human-centric service robot domain. We especially demonstrate how \hst\ can leverage and model interactions between humans consistently over multiple possible futures. Further, we show that our approach is cross-domain compatible with unconstrained outdoor pedestrian prediction, where data from a surveillance camera is used to predict pedestrians in different outdoor squares. Finally, we demonstrate how \hst\ can leverage vision-based features in human-centric environments to improve prediction accuracy, specifically in short history situations where prediction errors are high.

\smallskip \noindent \textbf{Datasets.}
To effectively investigate the performance of \hst\ in human-centric environments and the possible benefits that a detailed 3D skeletal representation of the human body can have, a dataset should (\rom{1})~include a \textit{diverse} range of indoor and outdoor environments, (\rom{2}) capture humans' movement from a \textit{robot} viewpoint (\rom{3}) in a natural \textit{unscripted} environment%
, (\rom{4})~provide labels for the position and skeletal keypoints ($\mathbf{k}$) of all agents at all timesteps, and (\rom{5})~be sufficiently \textit{large} to prevent over-fitting.
\begin{table}
\centering
\scriptsize
\setlength{\tabcolsep}{3pt}
\caption{A summary of different prediction datasets indicating our desiderata addressed by each dataset. Parentheses indicate that a desiderata is only partially fulfilled.}
\begin{tabular}{@{}l|ccccc@{}}
\toprule
     & \rom{1} Diverse & \rom{2} Robot &  \rom{3} Unscripted  & \rom{4} $\mathbf{k}$ & \rom{5} Large \\ \midrule
ETH~\cite{pellegrini2009you} \& UCY~\cite{lerner2007crowds}     & \xmark & \xmark & \cmark & (\xmark) & \xmark \\
AD~\cite{Ettinger_2021_ICCV, caesar2020nuscenes, Argoverse2, houston2021one}     & \xmark & \cmark & \cmark & (\xmark) & \cmark \\
Motion Capture~\cite{h36m_pami, AMASS:ICCV:2019}& \xmark & \xmark & \xmark & (\cmark) & (\xmark) \\
3DPW~\cite{vonMarcard2018}   & \xmark & \xmark & \xmark & \cmark & \xmark \\ \midrule
Adapted JRDB & \cmark & \cmark & \cmark & (\cmark) & \xmark \\ \bottomrule
\end{tabular}
\label{tab:datasets}
\end{table}
The evaluation of different datasets for human trajectory prediction in \cref{tab:datasets} takes into account the satisfaction 
of these requirements. Many of existing datasets are collected from a single top-down camera in a limited number of environments, such as the ETH~\cite{pellegrini2009you} and UCY~\cite{lerner2007crowds} pedestrian datasets. Others are specific to the autonomous driving domain~\cite{Ettinger_2021_ICCV, caesar2020nuscenes, Argoverse2, houston2021one}. While none of these datasets provide labels for skeleton keypoints, other datasets such as H3.6.M~\cite{h36m_pami}, AMASS~\cite{AMASS:ICCV:2019}, and 3DPW~\cite{vonMarcard2018} which are collected using a motion capture system or wearable IMU devices~\cite{vonMarcard2018}, do offer such labels. However, these datasets are limited to artificial environments and often feature stationary or scripted motions. %
Finally, while all these datasets provide labels for position, these labels are often hand labeled ground truth not representing noisy input data a robot would experience in the real-world during inference.

One dataset which is recorded in diverse human-centric environments using sensors (2 x 16 Channel LIDAR, 5 x Stereo RGB Cameras) on a mobile robotic platform is the JackRabbot Dataset and Benchmark (JRDB)~\cite{martin2021jrdb}. %
However, JRDB was created as a detection and tracking dataset rather than a prediction dataset. 
To make the data suitable for a prediction task, we first extract the robot motion from the raw sensor data to account for the robot's motion. Tracks are generated for both train and test split using the JRMOT~\cite{shenoi2020jrmot} detector and tracker. The ground truth labeled bounding-boxes on the train set were disregarded as they were exposed to filtering during the labeling process to the point where the smoothness eases the prediction task. 
We were able to increase the number of human tracks for training by associating the JRMOT detections to ground truth track labels via Hungarian matching, while on the test split we solely use JRMOT predictions. %

Due to factors such as distance, lighting and occlusion the pre-trained 3D pose estimator model (\cref{sec:model}) is not guaranteed to produce keypoints for all agents at all timesteps. We observed human keypoints information in $\sim 50\%$ of all timesteps for all agents in a distance of up to 7 meters from the robot.
The available data is split using $50\%$ of each scene as training data, $20\%$ as validation data and $30\%$ as test data.
We sub-sample the dataset from $15\si{Hz}$ to $3\si{Hz}$ keeping all of the intermediate samples and therefore increasing the number of datapoints by a factor of five.

In addition, we also compare our model to the ETH~\cite{pellegrini2009you} and UCY~\cite{lerner2007crowds} datasets. These are standard benchmarks for pedestrian trajectory prediction and enable a fair comparison of our architecture against other methods. 

\smallskip \noindent \textbf{Metrics.}
In consistency with prior work \cite{salzmann2020trajectron++, yuan2021agentformer, sadeghian2019sophie, mangalam2020not}, prediction quality is evaluated using the following metrics:

\begin{enumerate}[leftmargin=*]
    \item Minimum Average Displacement Error (\textit{minADE}): Minimum Mean $l_2$ distance between the ground truth and all $M$ future mode trajectories.
    
    \item Minimum Final Displacement Error (\textit{minFDE}): Minimum $l_2$ distance between the final positions of the ground truth and all $M$ future mode trajectories.
    
    \item Maximum Likelihood Average Displacement Error (\textit{MLADE}): Mean $l_2$ distance between the ground truth and the most likely mode trajectory.
    
    \item Negative Log Likelihood (\textit{NLL}) of the ground truth under the full parametric output distribution.
\end{enumerate}
Lower is better for all metrics.

\smallskip \noindent \textbf{Baselines.}
We re-implement the autonomous driving Scene Transformer architecture~\cite{ngiam2021scene}, where we match the number of Transformer layers to our architecture. We further compare to trajectory prediction architectures Trajectron++~\cite{salzmann2020trajectron++}, Agentformer~\cite{yuan2021agentformer}, SoPhie~\cite{sadeghian2019sophie}, and PECNet~\cite{mangalam2020not}.

\smallskip \noindent \textbf{Evaluation Protocol.} For the JRDB prediction dataset we report \textit{minADE}, \textit{MLADE}, and \textit{NLL} for 128k scene snippets from the test split including partially occluded agents. We use up to \SI{2}{\second} of history as input and predict the next \SI{4}{\second} future of the scene. 
Note that if the number of agents exceeds $N_{\text{max}} = 16$, we randomly select one agent and only consider and predict it and the $N_{\text{max}} -1 $ closest agents.
On the ETH and UCY dataset we follow the standard procedure to train in a leave-one-out fashion and evaluate \textit{minADE} and \textit{minFDE} on $20$ trajectories over a prediction horizon of 12 timesteps ($4.8\si{s}$) using 8 historic timesteps as input. We match the evaluation protocol of AgentFormer~\cite{yuan2021agentformer} by setting the number of modes to $M=20$. %

\begin{figure*}[b]
    \setlength{\fboxsep}{0pt}
    \centering
    \begin{subfigure}[t]{0.5\linewidth}
        \begin{subfigure}[t]{0.3136\linewidth}
            \centering
            \fbox{\includegraphics[width=\linewidth]{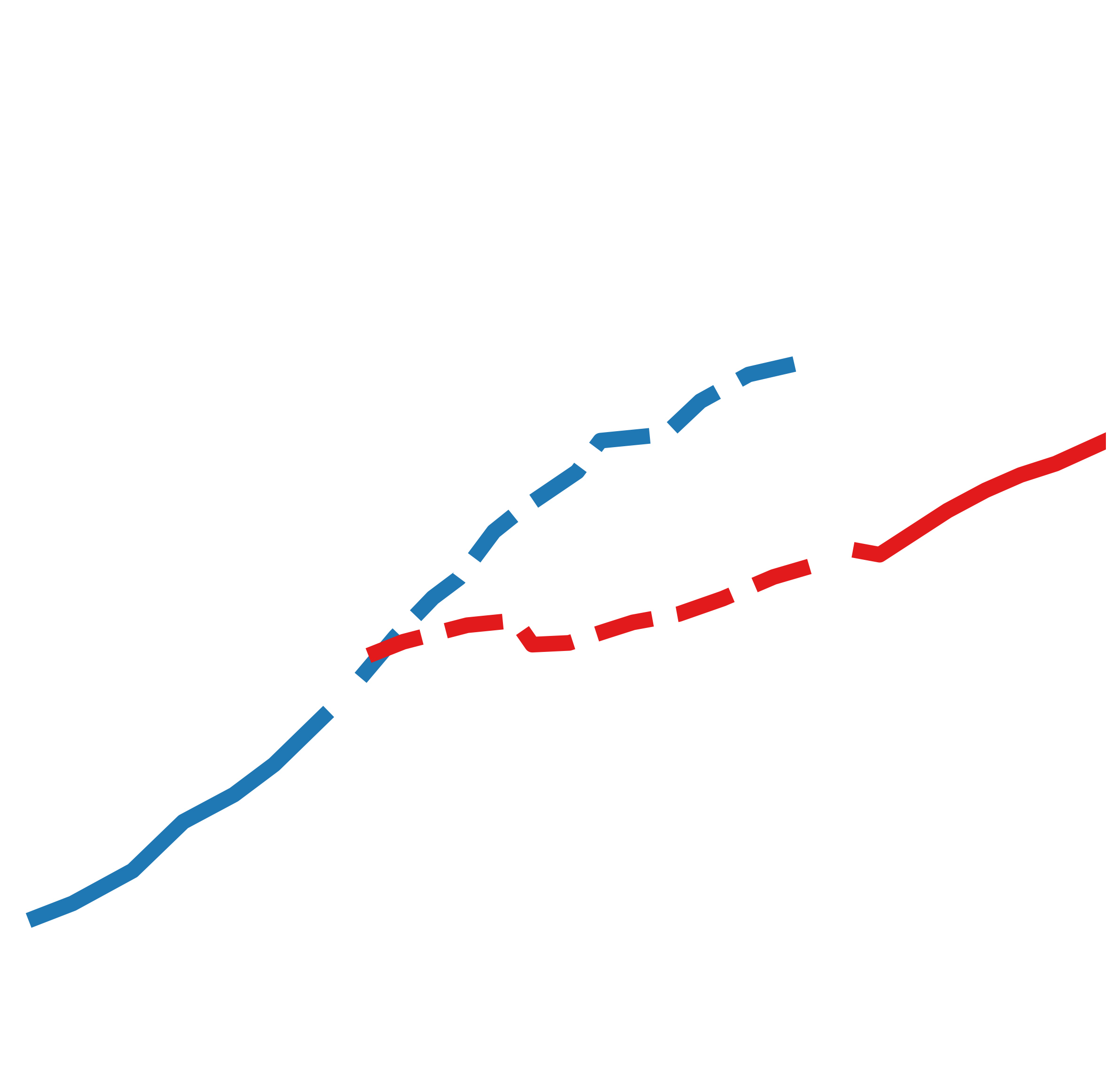}}
            \caption{}
        \end{subfigure}\hfill \vline \hfill
        \begin{subfigure}[t]{0.64\linewidth}
            \centering
            \begin{subfigure}[t]{0.49\linewidth}
                \centering
                \fbox{\includegraphics[width=\linewidth]{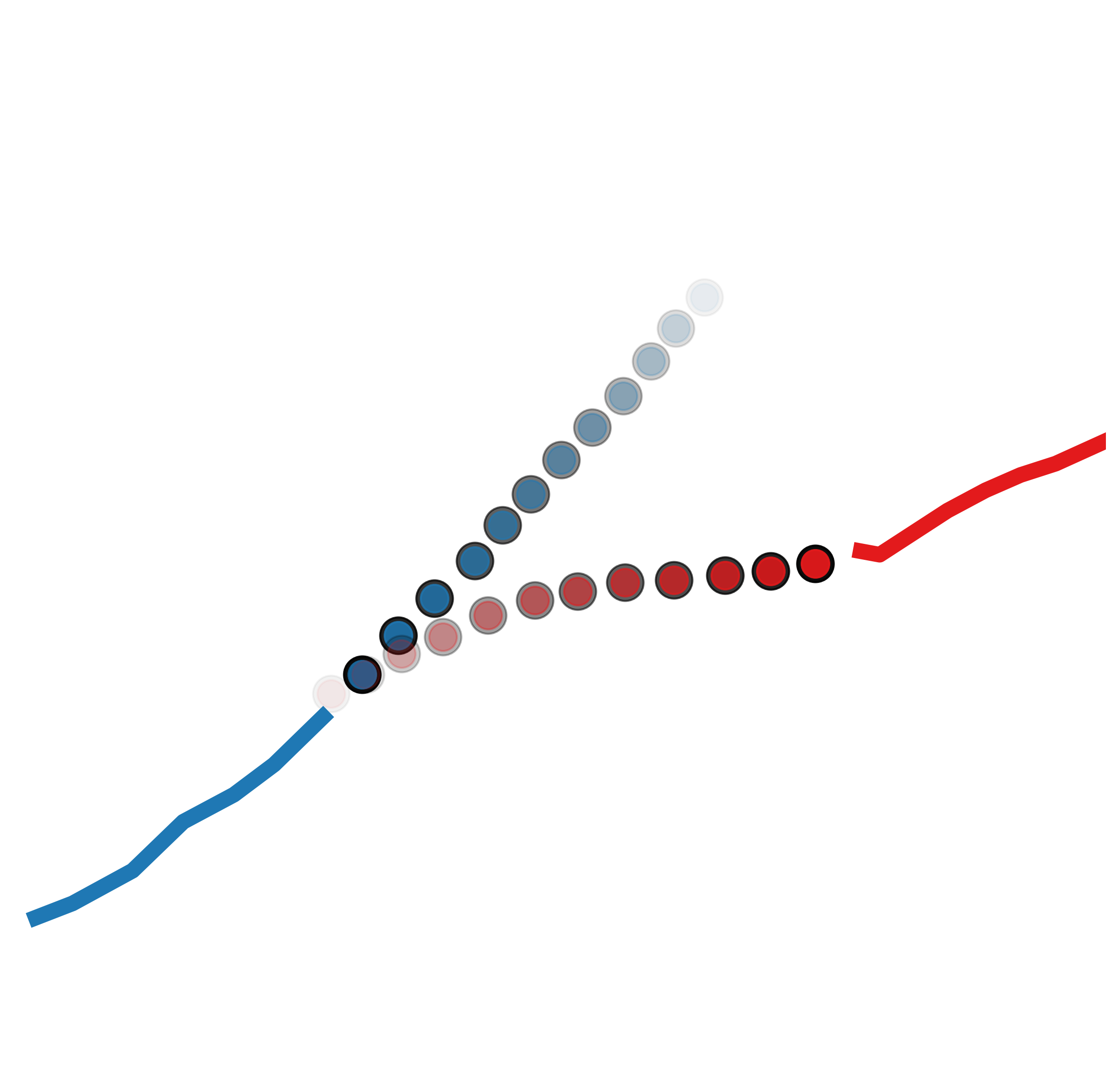}}
            \end{subfigure}\hfill
            \begin{subfigure}[t]{0.49\linewidth}
                \centering
                \fbox{\includegraphics[width=\linewidth]{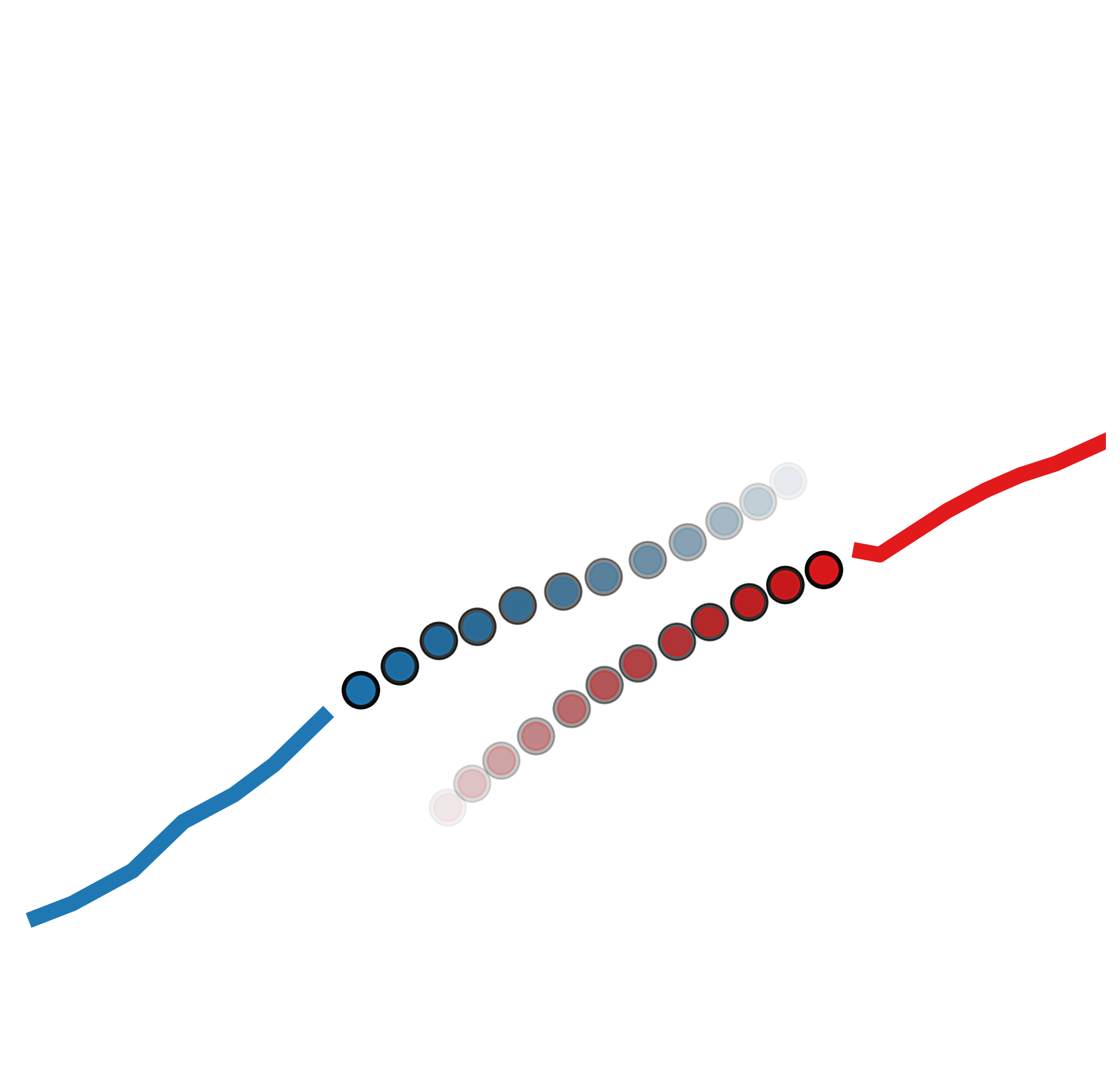}}
            \end{subfigure}
            \caption{}
        \end{subfigure}
    \end{subfigure}\hspace{4em}
    \begin{subfigure}[t]{0.321\linewidth}
            \centering
            \begin{subfigure}[t]{0.49\linewidth}
                \centering
                \fbox{\includegraphics[width=\linewidth]{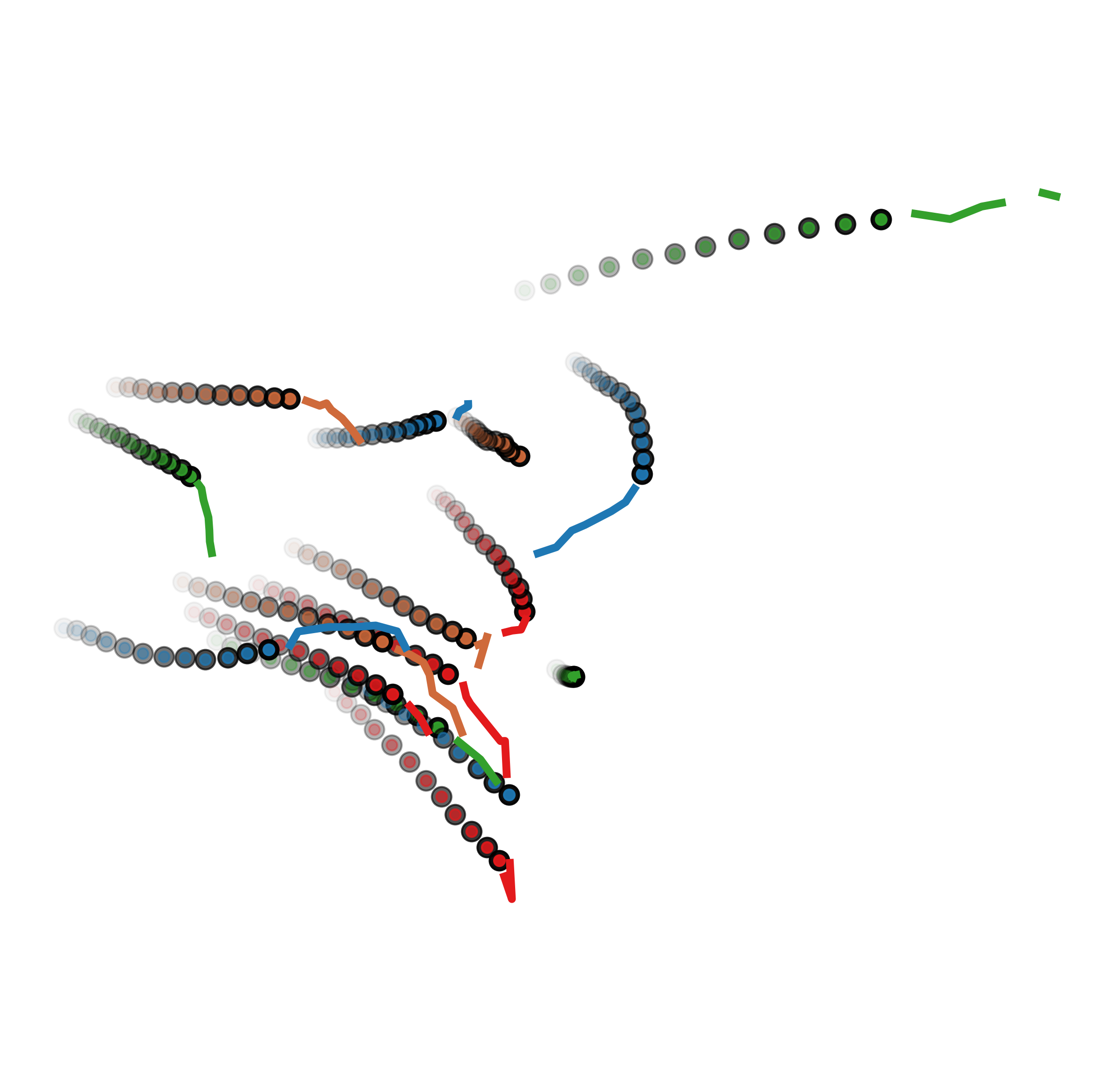}}
            \end{subfigure}\hfill
            \begin{subfigure}[t]{0.49\linewidth}
                \centering
                \fbox{\includegraphics[width=\linewidth]{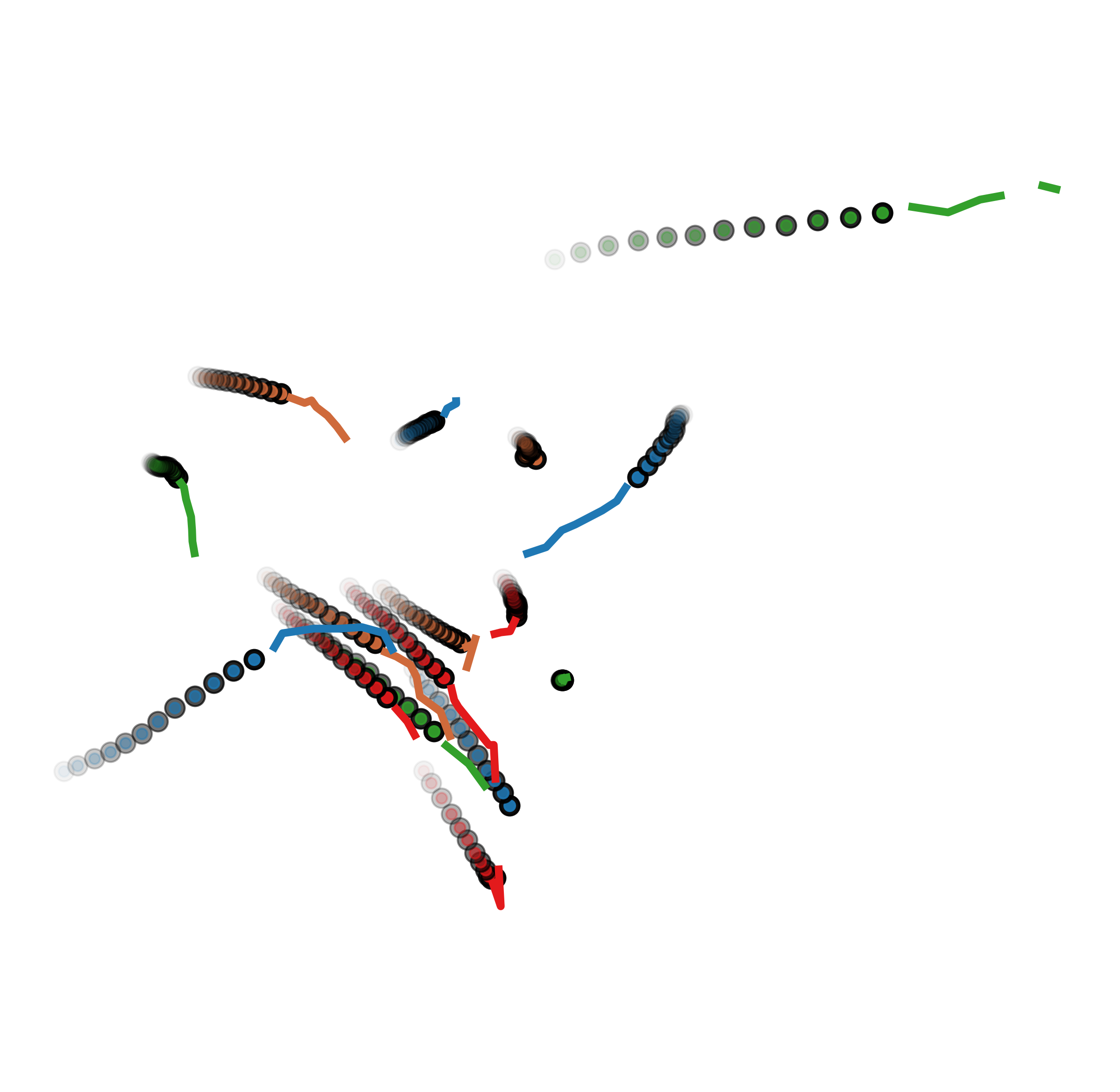}}
            \end{subfigure}
            \caption{}
        \end{subfigure}
    \caption{\textbf{Consistently modeled interactions in different predicted futures for a single scene in the x-y-plane.} Two humans approaching each other head on. (a) History (solid) and ground truth future (dashed) of both humans. (b) Two of the $M$ predicted futures (dots - increasing transparency with time) of the scene by \hst. 
    Within each mode the influence and reaction of both agents is consistent and reasonable. The humans' futures are predicted without collisions giving each other space to navigate within the specific predicted future mode of the scene. (c) Two predicted futures of a crowded scene.}%
    \label{fig:qual_modes}
\end{figure*}

\subsection{Trajectory Prediction in Human-centric Environments}\label{sec:exp_jrdb}

In \cref{tab:results_jrdb} and \cref{fig:qual_modes} we show quantitative and qualitative results of \hst's predictions in the human-centric environment. We show that in crowded human-centric environments the influence of interaction between humans has large benefits on the prediction accuracy of each individual. To show this, we compare against a version of our model which is trained to predict a single human at a time ignoring interactions with other agents. Subsequently, adding our full self-attention via self-alignment mechanism additionally increases the model's ability to capture interactions across time, leading to improvements across all metrics.
The capability to account for interactions between humans is qualitatively demonstrated in \cref{fig:qual_modes} where we show multiple predicted futures for a scene of two interacting humans. The two humans approach each other head on. The possible interactions to avoid collisions are modeled \emph{consistently} within each future. %

\begin{table}[t]
\centering
\scriptsize
\caption{\textbf{Comparison against Scene Transformer on JRDB prediction dataset.} \hst\ outperforms the original Scene Transformer on all metrics. Ablation shows that the interaction attention to other agents improves performance by comparing to a model predicting a single human at time. We also show the positive impact of Full Self-Attention.}
\renewcommand\tabcolsep{3pt}
\begin{tabular}{@{}lcc|cccccc@{}}
\toprule
\multicolumn{3}{c}{Model Configuration} & \textit{minADE} & \textit{MLADE} &  \textit{NLL}  \\ \midrule
\multicolumn{3}{l|}{Scene Transformer~\cite{ngiam2021scene}}  & $0.53$ & $0.86$ & $0.25$ \\ \midrule
& Full Self-Attention & Interaction Attention \\
\hst & \xmark & \xmark & $0.57$ & $0.93$ & $0.89$ \\
\hst & \xmark  & \cmark & $0.50$ & $0.84$ & $-0.02$ \\
\hst & \cmark & \cmark &  $\mathbf{0.48}$ & $\mathbf{0.80}$ & $\mathbf{-0.13}$ \\
\end{tabular}
\label{tab:results_jrdb}
\end{table}

\subsection{Vision-based Features}
In this section, we consider the adversarial setting, where the robot encounters a human unexpectedly, i.e., the robot observes a new human with little historical observations.
Prediction architectures solely relying on historic position information struggle in scenarios where no or only a limited amount of history of the human position is available to the model. Specifically, at the first instance of human detection, the experimental error is 200\% higher compared to full historic information over \SI{2}{\second}. Given the specifics of our targeted human-centric environment, where we are mostly interested in humans close to the robot, we are likely able to extract vision-based features for the human in addition to the position. Specifically, we revisit our research question: \emph{``Can information from human visual features lead to improved prediction accuracy?''}

\begin{figure}
    \begin{subfigure}[t]{\linewidth}
        \centering
        \fbox{\includegraphics[width=0.98\linewidth, trim={0.2cm 4.2cm 0.2cm 3.5cm},clip]{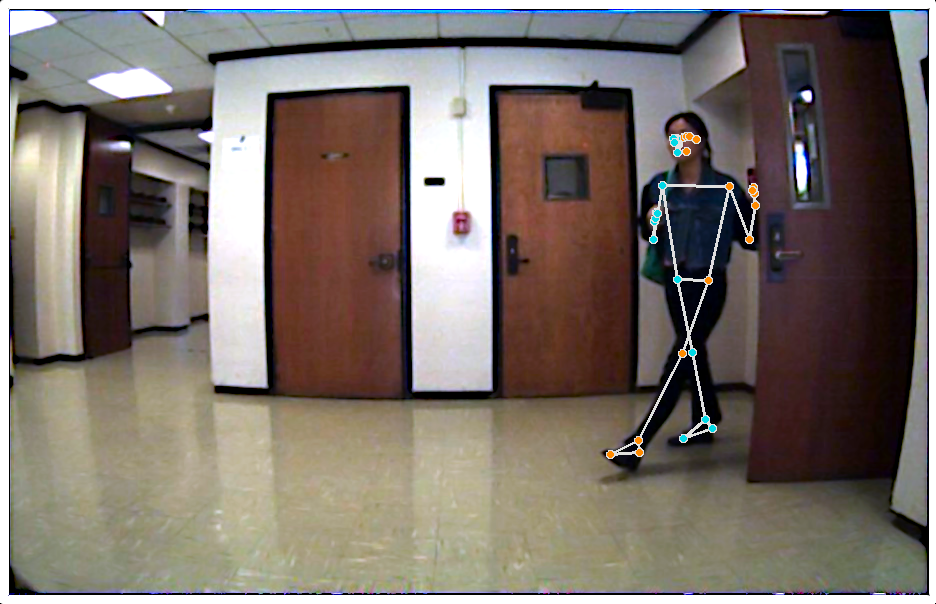}}
        \caption{First detection of person entering the scene.}
    \end{subfigure}\vspace{2pt}\\
    \begin{subfigure}[]{0.468\linewidth}
        \fbox{\includegraphics[width=\linewidth, trim={0cm 3cm 0cm 0.4cm},clip]{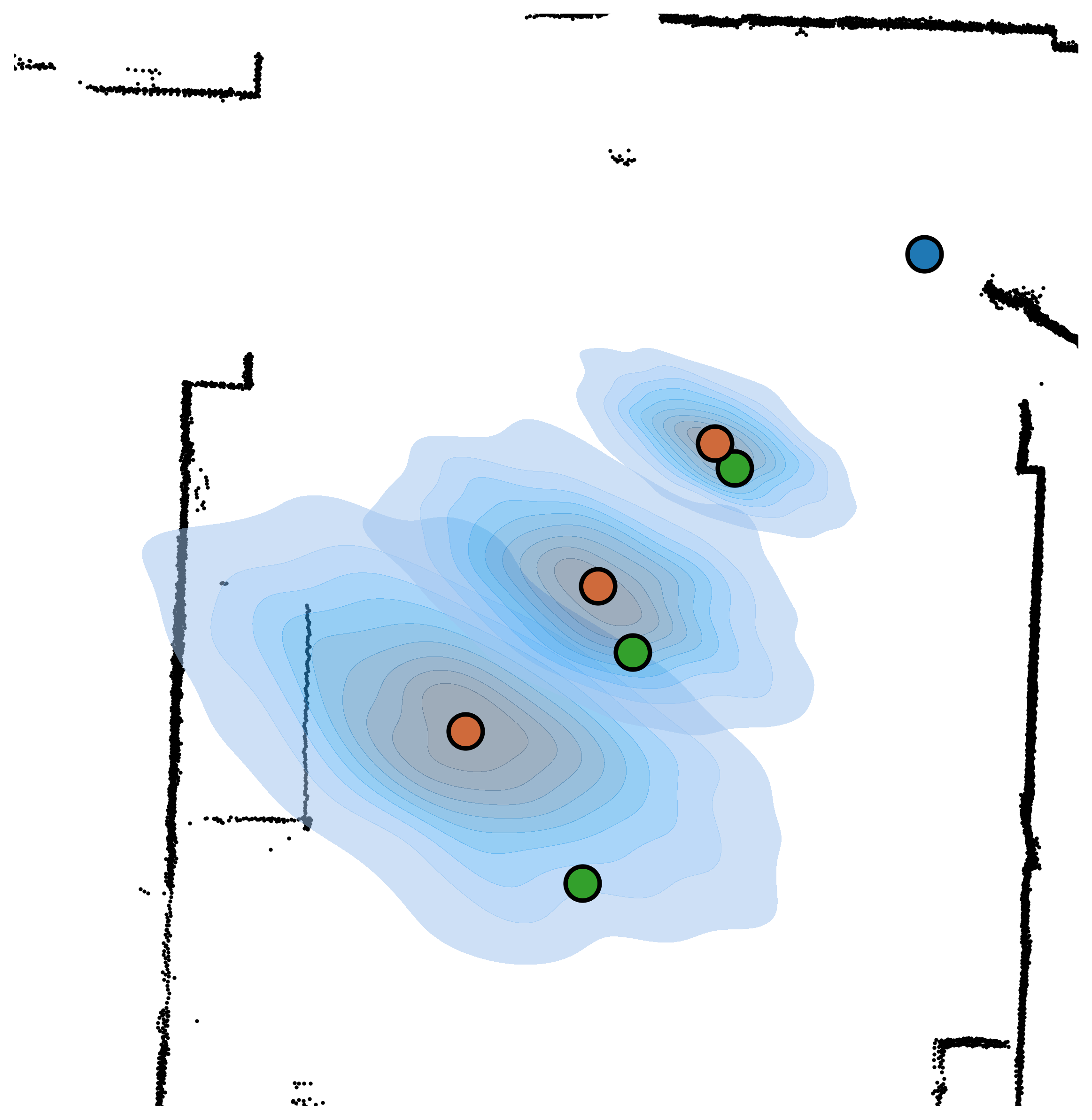}}
        \caption{Prediction with keypoints.}
    \end{subfigure}\hspace{10pt}%
    \begin{subfigure}[]{0.468\linewidth}
        \fbox{\includegraphics[width=\linewidth, trim={0cm 3cm 0cm 0.4cm},clip]{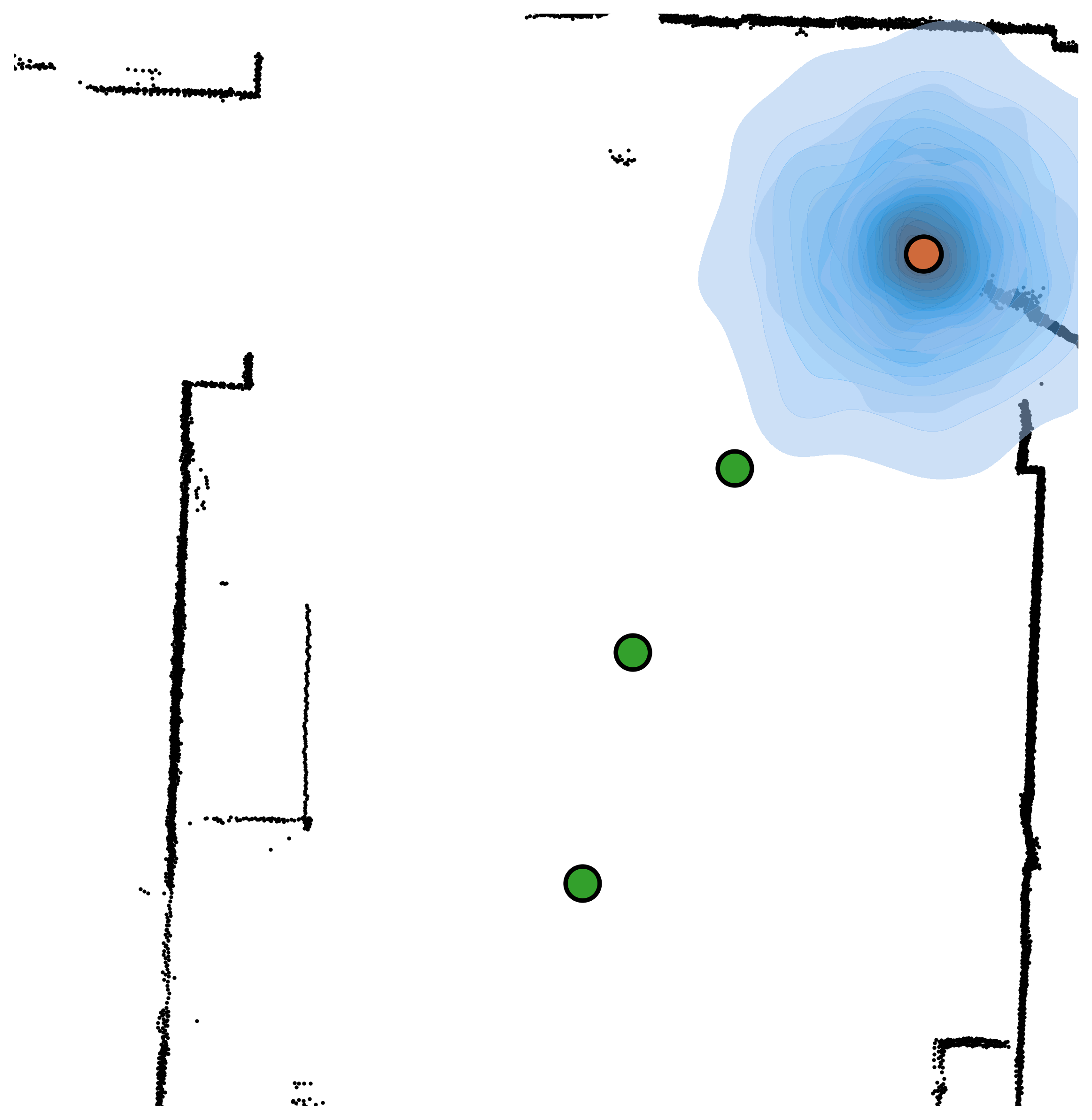}}
        \caption{Prediction without keypoints.}
    \end{subfigure}%
    \caption{A visualization of the predicted trajectory distributions for a new human agent entering the scene through the door on the right as viewed in (a). For \emph{both} (b) and (c) the \hst\ model does not have any historic information here and only has access to the current frame. The plot of future trajectory distributions in (b) and (c) show the effect of using and not using skeletal keypoints (respectively) as input in that single frame. Without pose keypoints the \hst\ model predicts the agent to be most-likely stationary while, with keypoints as input, it can reason that the human is moving and correctly anticipates the direction.
    Blue dot is the detected human at the initial frame, orange dots are our most likely mode predictions with corresponding distribution shown with blue shading, and green dots are the ground truth human future (actually executed trajectory by the human).}
    \label{fig:qual}
\end{figure}

Before answering this question quantitatively we show a clarifying visual example in \cref{fig:qual} where a human just entered the scene through a door and is first detected. When solely relying on historic position information the most likely prediction by the model is a stationary agent. However, when we employ the pre-trained skeleton keypoints estimator to provide pose keypoints as additional input to our model the model correctly realizes if the human is in a walking motion and how the human is oriented, accurately predicting the most likely future trajectory.

\begin{figure}
    \centering
    \includegraphics[width=0.85\linewidth]{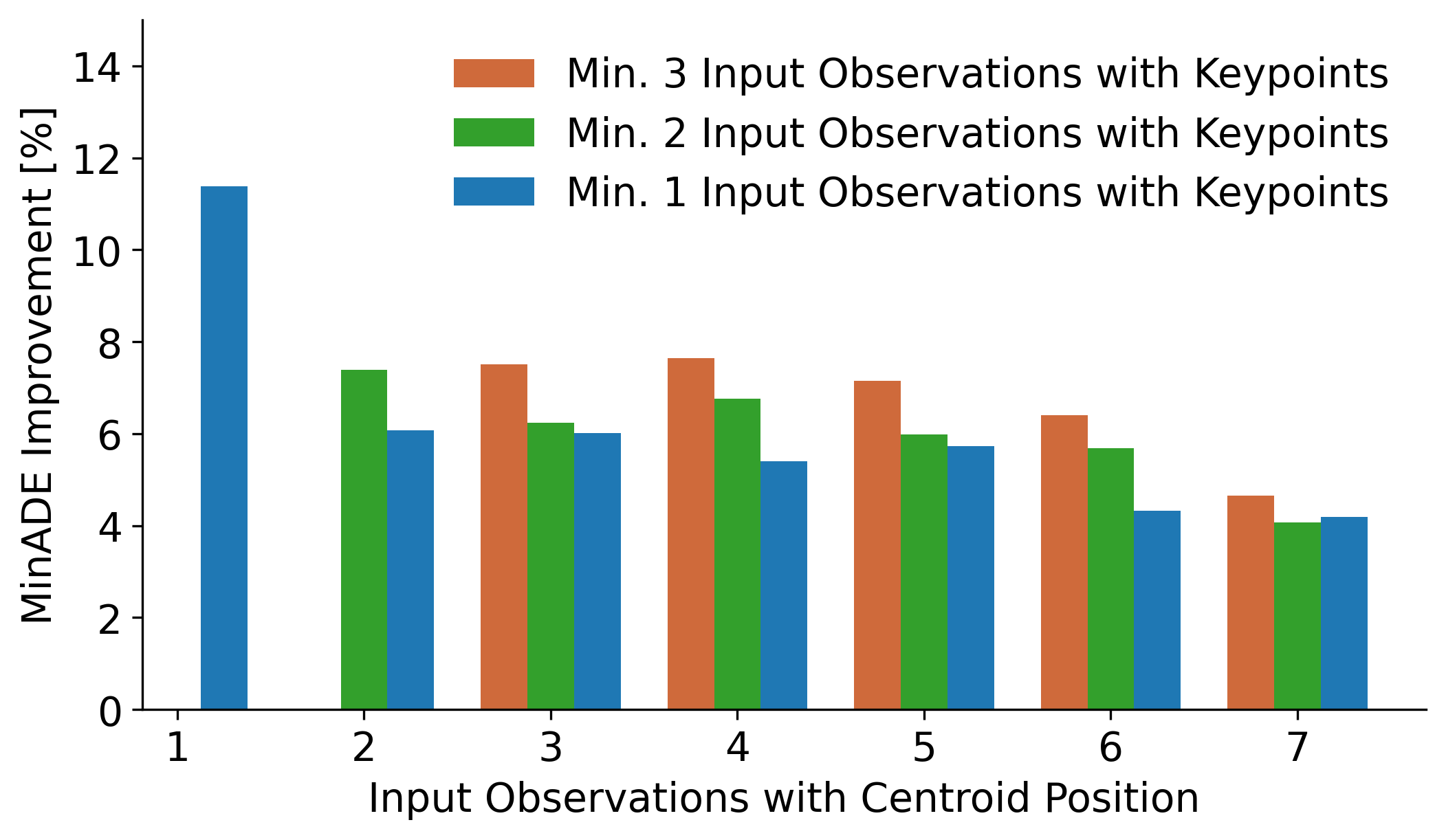}
    \caption{Impact of vision-based features conditioned on different number of consecutive non occluded input timesteps. }
    \label{fig:vision_imp}
\end{figure}

Quantitatively, during evaluation, when keypoints are available on the first detection we observe a substantial prediction improvement of up to 11\% (\cref{fig:vision_imp}). When additional timesteps with position information are available the improvement using keypoints vs not using keypoints averages between 5\% and 10\%. The relative improvement generally increases with the number of timesteps with keypoints in the history and decreases with the number of historic position information.

Finally, we want to provide an outlook of how much vision-based features can improve prediction performance if available for all agents at all timesteps. We therefore enforce feature parity between position and skeletal keypoints features by disregarding position information without associated keypoints (\cref{tab:results_parity}). We find that a relative improvement of around 10\% is achievable using our in-the-wild vision-based features. The baseline is naturally worse than in \cref{tab:results_jrdb} as we partially disregard historic position information.

\begin{table}
\centering
\scriptsize
\setlength{\tabcolsep}{4pt}
\caption{Vision-based features relatively improve prediction across all metrics on a prediction horizon of \SI{4}{\second}. To create feature parity between position and keypoints features we ignore all position history without detected skeletal keypoints during training and evaluation.}
\begin{tabular}{@{}l|cc|cc|cc|cc@{}}
\toprule
  & \multicolumn{2}{c}{\textit{minADE}} &  \multicolumn{2}{c}{\textit{NLL}} & \multicolumn{2}{c}{\textit{minADE} @ \SI{2}{\second}} & \multicolumn{2}{c}{\textit{minADE} @ \SI{4}{\second}} \\ \midrule
Baseline & $0.56$ & $0\%$  & $1.04$ & $0\%$ & $0.46$ & $0\%$ & $0.87$ & $0\%$ \\
Head Orient. & $0.53$ & $-5\%$  & $0.89$ & $-14\%$ & $0.42$ & $-9\%$ & $0.79$ & $-9\%$\\
Keypoints & $\mathbf{0.51}$ & $-9\%$  & $\mathbf{0.85}$ & $-18\%$ & $\mathbf{0.40}$ & $-13\%$ & $\mathbf{0.77}$ & $-11\%$
\end{tabular}
\label{tab:results_parity}
\end{table}

\subsection{Pedestrian Dataset}
In addition to showing \hst's capabilities in a robotics specific environment we will further validate our architecture against a range of state-of-the-art prediction methods.  For this we use ETH/UCY which is prolific in the trajectory prediction community while also being the dataset which we think is the closest to the human-centric environments that we would like to explore: On the ETH and UCY dataset, we either improve current state-of-the-art methods or we are on par with them on 4 out of the 5 scenes. %

\begin{table*}[ht]
\centering
\scriptsize
\setlength{\tabcolsep}{17pt}
\caption{Results on ETH and UCY dataset. Our method is the best on four of five subsets.}
\begin{tabular}{@{}l|cccccc@{}}
\toprule
\multirow{2}{*}{Method}  & \multicolumn{6}{c}{\textit{minADE}$_{20}$ / \textit{minFDE}$_{20}$}  \\
  & ETH & Hotel & Univ & Zara1 & Zara2 & Average  \\ \midrule
SoPhie~\cite{sadeghian2019sophie} & $0.70\;/\;1.43$ & $0.76\;/\;1.67$ & $0.54\;/\;1.24$ & $0.30\;/\;0.63$ & $0.38\;/\;0.78$ & $0.54\;/\;1.15$ \\
PECNet~\cite{mangalam2020not} & $0.54\;/\;0.87$ & $0.18\;/\;0.24$ & $0.35\;/\;0.60$ & $0.22\;/\;0.39$ & $0.17\;/\;0.30$ & $0.29\;/\;0.48$ \\
Trajectron++~\cite{salzmann2020trajectron++} & $0.43\;/\;0.86$ & $0.12\;/\;0.19$ & $\mathbf{0.22}\;/\;0.44$ & $\mathbf{0.17}\;/\;0.32$ & $\mathbf{0.11}\;/\;0.25$ & $\mathbf{0.21}\;/\;0.41$ \\
AgentFormer~\cite{yuan2021agentformer}\textsuperscript{1} & $0.45\;/\;0.75$ & $0.14\;/\;0.22$ & $0.25\;/\;0.45$ & $0.18\;/\;0.30$ & $0.14\;/\;\mathbf{0.24}$ & $0.23\;/\;0.39$ \\ 
Scene Transformer~\cite{ngiam2021scene} & $0.50\;/\;0.76$ & $0.14\;/\;0.20$ & $0.29\;/\;\mathbf{0.42}$ & $0.22\;/\;0.36$ & $0.16\;/\;0.27$ & $0.31\;/\;0.40$ \\ \midrule
\hst & $\mathbf{0.41}\;/\;\mathbf{0.73}$ & $\mathbf{0.10}\;/\;\mathbf{0.14}$ & $0.24\;/\;0.44$ & $\mathbf{0.17}\;/\;\mathbf{0.30}$ & $0.14\;/\;\mathbf{0.24}$ & $\mathbf{0.21}\;/\;\mathbf{0.37}$\\
\end{tabular}
\label{tab:results_eth}
\end{table*}\vspace{-1.0em}

\section{Discussion}\label{sec:discussion}
Simply representing a human as its spatial position, as commonly done in autonomous driving environments, does deliver a baseline prediction performance in human-centric service robot environments. However, it suffers in challenging settings where the history of a human is limited. Specifically in these situations we demonstrate how the Human Scene Transformer can leverage vision-based features to improve prediction accuracy (\cref{fig:qual}). Beyond scenarios such as when robot and humans encounter each other in blind corners, general improvement trends using in-the-wild skeletal pose detections were also observed with more observations as shown in \cref{fig:vision_imp}. Another intuitive assumption which we can support quantitatively in \cref{tab:results_parity} is that the full skeletal keypoints (full human pose) hold additional information over just the head orientation (where is the human looking). This is expected as the head orientation can give away a possible direction for the trajectory, while the full keypoints can be more informative about the speed the human is approaching, e.g., running or slowly walking.

\cref{fig:qual_modes} and \cref{tab:results_jrdb} demonstrate \hst's capability to model consistent interactions between agents and use these influences to improve the overall prediction substantially. This is especially useful for the crowded spaces a service robot navigates and opens opportunities not just modeling human-to-human interactions but robot-human interactions.

\smallskip\noindent \textbf{Limitations.}
Exploring this new domain of human-centric environments we recognize that numerous limitations exist: 
We showed that an on par relationship between positions and keypoint features leads to relative improvements and expect the performance to be similarly correlated to the quality of the detected 3D keypoints. We therefore think that an 3D skeletal keypoints estimator, specifically designed for robotic applications, increasing both the number of successful detections and the quality of each detected skeleton would improve performance for the prediction task as well as other robot tasks in close human contact (e.g. handover). Such an estimator could make use of the full sensor suite of a robotic platform, fusing camera and LiDAR information. We are happy to see first works~\cite{zanfir2022hum3dil} in this direction and hope that our findings encourage the research community to pursue this path within the domain of human-centric environments.

In \cref{sec:model} we presented a way to include scene context via a point cloud but were unable to see any predictive benefit from this information. We attribute this to three factors: 
the small size of the adapted JRDB dataset, the limited number of locations (29) in which the data was recorded, and the raw point lack semantic labels, e.g. a door is indistinguishable from a wall. These factors highlight the potential for further work towards better representations of point clouds.

\addtocounter{footnote}{1}
\footnotetext{We report publicly available updated numbers for AgentFormer (\href{https://arxiv.org/pdf/2103.14023.pdf}{arxiv}/\href{https://github.com/Khrylx/AgentFormer/issues/5}{github}) which differ from the original publication.} %

\smallskip \noindent \textbf{Conclusion.}
This work introduced the task of human trajectory prediction into the domain of human-centric service robots. We demonstrated that the proximity of robot and humans in such environments can be leveraged to improve prediction performance by explicitly incorporating vision-based human features. We showed HST achieves strong prediction improvements using in-the-wild 3D pose representations for the critical situation of agents being first detected in close proximity to the robot.
A quantitative ablation using paired position and vision-based features highlighted the influence of different visual features with skeletal keypoints providing the highest gains across all metrics.
Our Transformer based architecture is flexible in accommodating different feature inputs while also achieving state-of-the-art results on a common pedestrian prediction dataset without visual features outside the domain of human-centric service robot environments. We think that the unstructured and uncompressed nature of environment point-clouds fits nicely with the permutation invariance property of Transformer architectures and are therefore excited to further explore this direction in the future. We hope that our findings will inspire further research in the human-centric domain and in developing improved methods for generating accurate 3D vision-based human representation for service robotics applications.

\vspace{-0.5em}
{
\scriptsize
\setlength{\biblabelsep}{0.35em} 
\setlength{\leftmargin}{50em}
\printbibliography

@INPROCEEDINGS {Argoverse2,
  author = {Benjamin Wilson and William Qi and Tanmay Agarwal and John Lambert and Jagjeet Singh and Siddhesh Khandelwal and Bowen Pan and Ratnesh Kumar and Andrew Hartnett and Jhony Kaesemodel Pontes and Deva Ramanan and Peter Carr and James Hays},
  title = {\href{https://openreview.net/pdf?id=vKQGe36av4k}{Argoverse 2: Next Generation Datasets for Self-driving Perception and Forecasting}},
  booktitle = {NeurIPS Datasets and Benchmarks},
  year = {2021}
}

@article{mahdavian2022stpotr,
  title={\href{https://arxiv.org/pdf/2209.07600.pdf}{STPOTR: Simultaneous Human Trajectory and Pose Prediction Using a Non-Autoregressive Transformer for Robot Following Ahead}},
  author={Mahdavian, Mohammad and Nikdel, Payam and TaherAhmadi, Mahdi and Chen, Mo},
  journal={arXiv:2209.07600},
  year={2022}
}

@inproceedings{salzmann2022motron,
  title={\href{https://openaccess.thecvf.com/content/CVPR2022/papers/Salzmann_Motron_Multimodal_Probabilistic_Human_Motion_Forecasting_CVPR_2022_paper.pdf}{Motron: Multimodal Probabilistic Human Motion Forecasting}},
  author={Salzmann, Tim and Pavone, Marco and Ryll, Markus},
  booktitle={Conference on Computer Vision and Pattern Recognition},
  pages={6457--6466},
  year={2022}
}

@inproceedings{salzmann2020trajectron++,
  title={\href{https://www.ecva.net/papers/eccv_2020/papers_ECCV/papers/123630664.pdf}{Trajectron++: Dynamically-feasible trajectory forecasting with heterogeneous data}},
  author={Salzmann, Tim and Ivanovic, Boris and Chakravarty, Punarjay and Pavone, Marco},
  booktitle={European Conference on Computer Vision},
  pages={683--700},
  year={2020},
  organization={Springer}
}

@inproceedings{ngiam2021scene,
  author    = {Jiquan Ngiam and
               Vijay Vasudevan and
               Benjamin Caine and
               Zhengdong Zhang and
               Hao{-}Tien Lewis Chiang and
               Jeffrey Ling and
               Rebecca Roelofs and
               Alex Bewley and
               Chenxi Liu and
               Ashish Venugopal and
               David J. Weiss and
               Ben Sapp and
               Zhifeng Chen and
               Jonathon Shlens},
  title     = {\href{https://openreview.net/pdf?id=Wm3EA5OlHsG}{Scene Transformer: {A} unified architecture for predicting future
               trajectories of multiple agents}},
  booktitle = {International Conference on Learning Representations},
  publisher = {OpenReview.net},
  year      = {2022},
  url       = {https://openreview.net/forum?id=Wm3EA5OlHsG},
  bibsource = {dblp computer science bibliography, https://dblp.org}
}

@inproceedings{adeli2021tripod,
  title={\href{https://openaccess.thecvf.com/content/ICCV2021/papers/Adeli_TRiPOD_Human_Trajectory_and_Pose_Dynamics_Forecasting_in_the_Wild_ICCV_2021_paper.pdf}{Tripod: Human trajectory and pose dynamics forecasting in the wild}},
  author={Adeli, Vida and Ehsanpour, Mahsa and Reid, Ian and Niebles, Juan Carlos and Savarese, Silvio and Adeli, Ehsan and Rezatofighi, Hamid},
  booktitle={International Conference on Computer Vision},
  pages={13390--13400},
  year={2021}
}

@article{chen2020pedestrian,
  title={\href{https://ieeexplore.ieee.org/document/9153745}{Pedestrian Trajectory Prediction in Heterogeneous Traffic Using Pose Keypoints-Based Convolutional Encoder-Decoder Network}},
  author={Chen, Kai and Song, Xiao and Ren, Xiaoxiang},
  journal={TCSVT},
  volume={31},
  number={5},
  pages={1764--1775},
  year={2020},
  publisher={IEEE}
}

@article{czech2022board,
  title={\href{https://arxiv.org/pdf/2210.11999.pdf}{On-Board Pedestrian Trajectory Prediction Using Behavioral Features}},
  author={Czech, Phillip and Braun, Markus and Kre{\ss}el, Ulrich and Yang, Bin},
  journal={arXiv:2210.11999},
  year={2022}
}

@article{vendrow2022somoformer,
  title={\href{https://arxiv.org/pdf/2208.14023.pdf}{SoMoFormer: Multi-Person Pose Forecasting with Transformers}},
  author={Vendrow, Edward and Kumar, Satyajit and Adeli, Ehsan and Rezatofighi, Hamid},
  journal={arXiv:2208.14023},
  year={2022}
}

@inproceedings{pellegrini2009you,
  title={\href{https://ieeexplore.ieee.org/document/5459260}{You'll never walk alone: Modeling social behavior for multi-target tracking}},
  author={Pellegrini, Stefano and Ess, Andreas and Schindler, Konrad and Van Gool, Luc},
  booktitle={International Conference on Computer Vision},
  pages={261--268},
  year={2009},
  organization={IEEE}
}

@inproceedings{lerner2007crowds,
  title={\href{https://onlinelibrary.wiley.com/doi/full/10.1111/j.1467-8659.2007.01089.x}{Crowds by example}},
  author={Lerner, Alon and Chrysanthou, Yiorgos and Lischinski, Dani},
  booktitle={Computer graphics forum},
  volume={26},
  pages={655--664},
  year={2007},
  organization={Wiley Online Library}
}

@InProceedings{Ettinger_2021_ICCV,
author={Ettinger, Scott and Cheng, Shuyang and Caine, Benjamin and Liu, Chenxi and Zhao, Hang and Pradhan, Sabeek and Chai, Yuning and Sapp, Ben and Qi, Charles R. and Zhou, Yin and Yang, Zoey and Chouard, Aur'elien and Sun, Pei and Ngiam, Jiquan and Vasudevan, Vijay and McCauley, Alexander and Shlens, Jonathon and Anguelov, Dragomir}, title={\href{https://openaccess.thecvf.com/content/ICCV2021/papers/Ettinger_Large_Scale_Interactive_Motion_Forecasting_for_Autonomous_Driving_The_Waymo_ICCV_2021_paper.pdf}{Large Scale Interactive Motion Forecasting for Autonomous Driving: The Waymo Open Motion Dataset}}, booktitle= {International Conference on Computer Vision}, year={2021}, pages={9710-9719} }

@inproceedings{caesar2020nuscenes,
  title={\href{https://openaccess.thecvf.com/content_CVPR_2020/papers/Caesar_nuScenes_A_Multimodal_Dataset_for_Autonomous_Driving_CVPR_2020_paper.pdf}{nuscenes: A multimodal dataset for autonomous driving}},
  author={Caesar, Holger and Bankiti, Varun and Lang, Alex H and Vora, Sourabh and Liong, Venice Erin and Xu, Qiang and Krishnan, Anush and Pan, Yu and Baldan, Giancarlo and Beijbom, Oscar},
  booktitle={Conference on Computer Vision and Pattern Recognition},
  pages={11621--11631},
  year={2020}
}

@inproceedings{houston2021one,
  title={\href{https://arxiv.org/pdf/2006.14480.pdf}{One thousand and one hours: Self-driving motion prediction dataset}},
  author={Houston, John and Zuidhof, Guido and Bergamini, Luca and Ye, Yawei and Chen, Long and Jain, Ashesh and Omari, Sammy and Iglovikov, Vladimir and Ondruska, Peter},
  booktitle={Conference on Robot Learning},
  pages={409--418},
  year={2021},
  organization={PMLR}
}

@article{h36m_pami,
author = {Ionescu, Catalin and Papava, Dragos and Olaru, Vlad and Sminchisescu,  Cristian},
title = {\href{https://ieeexplore.ieee.org/document/6682899}{Human3.6M: Large Scale Datasets and Predictive Methods for 3D Human Sensing in Natural Environments}},
journal = {IEEE Transactions on Pattern Analysis and Machine Intelligence},
publisher = {IEEE Computer Society},
volume = {36},
number = {7},
pages = {1325-1339},
year = {2014}
}

@conference{AMASS:ICCV:2019,
  title = {\href{https://www.computer.org/csdl/proceedings-article/iccv/2019/480300f441/1hVlFDUXIs0}{{AMASS}: Archive of Motion Capture as Surface Shapes}},
  author = {Mahmood, Naureen and Ghorbani, Nima and Troje, Nikolaus F. and Pons-Moll, Gerard and Black, Michael J.},
  booktitle = {International Conference on Computer Vision},
  pages = {5442--5451},
  month = oct,
  year = {2019},
  month_numeric = {10}
}

@inproceedings{vonMarcard2018,
title = {\href{https://openaccess.thecvf.com/content_ECCV_2018/papers/Timo_von_Marcard_Recovering_Accurate_3D_ECCV_2018_paper.pdf}{Recovering Accurate 3D Human Pose in The Wild Using IMUs and a Moving Camera}},
author = {von Marcard, Timo and Henschel, Roberto and Black, Michael and Rosenhahn, Bodo and Pons-Moll, Gerard},
booktitle = {European Conference on Computer Vision},
year = {2018}
}

@article{martin2021jrdb,
    title={\href{https://arxiv.org/pdf/1910.11792.pdf}{Jrdb: A dataset and benchmark of egocentric robot visual perception of humans in built environments}},
    author={Martin-Martin, Roberto and Patel, Mihir and Rezatofighi, Hamid and Shenoi, Abhijeet and Gwak, JunYoung and Frankel, Eric and Sadeghian, Amir and Savarese, Silvio},
    journal={IEEE Transactions on Pattern Analysis and Machine Intelligence},
    year={2021},
    publisher={IEEE}
}

@inproceedings{shenoi2020jrmot,
  title={\href{https://ras.papercept.net/images/temp/IROS/files/1436.pdf}{Jrmot: A real-time 3d multi-object tracker and a new large-scale dataset}},
  author={Shenoi, Abhijeet and Patel, Mihir and Gwak, JunYoung and Goebel, Patrick and Sadeghian, Amir and Rezatofighi, Hamid and Martin-Martin, Roberto and Savarese, Silvio},
  booktitle={International Conference on Intelligent Robots and Systems},
  pages={10335--10342},
  year={2020},
  organization={IEEE}
}

@article{grishchenko2022blazepose,
  title={\href{https://arxiv.org/pdf/2206.11678.pdf}{BlazePose GHUM Holistic: Real-time 3D Human Landmarks and Pose Estimation}},
  author={Grishchenko, Ivan and Bazarevsky, Valentin and Zanfir, Andrei and Bazavan, Eduard Gabriel and Zanfir, Mihai and Yee, Richard and Raveendran, Karthik and Zhdanovich, Matsvei and Grundmann, Matthias and Sminchisescu, Cristian},
  journal={Computer Vision for AR/VR},
  year={2022}
}

@inproceedings{yuan2021agentformer,
  title={\href{https://arxiv.org/pdf/2103.14023.pdf}{Agentformer: Agent-aware transformers for socio-temporal multi-agent forecasting}},
  author={Yuan, Ye and Weng, Xinshuo and Ou, Yanglan and Kitani, Kris M},
  booktitle={International Conference on Computer Vision},
  pages={9813--9823},
  year={2021}
}

@inproceedings{mangalam2021goals,
  title={\href{https://openaccess.thecvf.com/content/ICCV2021/papers/Mangalam_From_Goals_Waypoints__Paths_to_Long_Term_Human_Trajectory_ICCV_2021_paper.pdf}{From goals, waypoints \& paths to long term human trajectory forecasting}},
  author={Mangalam, Karttikeya and An, Yang and Girase, Harshayu and Malik, Jitendra},
  booktitle={International Conference on Computer Vision},
  pages={15233--15242},
  year={2021}
}

@inproceedings{ivanovic2019trajectron,
  title={\href{https://openaccess.thecvf.com/content_ICCV_2019/papers/Ivanovic_The_Trajectron_Probabilistic_Multi-Agent_Trajectory_Modeling_With_Dynamic_Spatiotemporal_Graphs_ICCV_2019_paper.pdf}{The trajectron: Probabilistic multi-agent trajectory modeling with dynamic spatiotemporal graphs}},
  author={Ivanovic, Boris and Pavone, Marco},
  booktitle={International Conference on Computer Vision},
  pages={2375--2384},
  year={2019}
}

@article{nayakanti2022wayformer,
  title={\href{https://arxiv.org/pdf/2207.05844.pdf}{Wayformer: Motion forecasting via simple \& efficient attention networks}},
  author={Nayakanti, Nigamaa and Al-Rfou, Rami and Zhou, Aurick and Goel, Kratarth and Refaat, Khaled S and Sapp, Benjamin},
  journal={arXiv:2207.05844},
  year={2022}
}

@article{zanfir2022hum3dil,
  title={\href{https://openreview.net/pdf?id=jTh3rdEF3LH}{HUM3DIL: Semi-supervised Multi-modal 3D Human Pose Estimation for Autonomous Driving}},
  author={Zanfir, Andrei and Zanfir, Mihai and Gorban, Alexander and Ji, Jingwei and Zhou, Yin and Anguelov, Dragomir and Sminchisescu, Cristian},
  journal={arXiv:2212.07729},
  year={2022}
}

@inproceedings{corona2020context,
  title={\href{https://openaccess.thecvf.com/content_CVPR_2020/papers/Corona_Context-Aware_Human_Motion_Prediction_CVPR_2020_paper.pdf}{Context-aware human motion prediction}},
  author={Corona, Enric and Pumarola, Albert and Alenya, Guillem and Moreno-Noguer, Francesc},
  booktitle={Conference on Computer Vision and Pattern Recognition},
  pages={6992--7001},
  year={2020}
}

@inproceedings{narayanan2020proxemo,
  title={\href{https://arxiv.org/pdf/2003.01062.pdf}{Proxemo: Gait-based emotion learning and multi-view proxemic fusion for socially-aware robot navigation}},
  author={Narayanan, Venkatraman and Manoghar, Bala Murali and Dorbala, Vishnu Sashank and Manocha, Dinesh and Bera, Aniket},
  booktitle={International Conference on Intelligent Robots and Systems},
  pages={8200--8207},
  year={2020},
  organization={IEEE}
}

@inproceedings{yagi2018future,
  title={\href{https://openaccess.thecvf.com/content_cvpr_2018/papers/Yagi_Future_Person_Localization_CVPR_2018_paper.pdf}{Future person localization in first-person videos}},
  author={Yagi, Takuma and Mangalam, Karttikeya and Yonetani, Ryo and Sato, Yoichi},
  booktitle={Conference on Computer Vision and Pattern Recognition},
  pages={7593--7602},
  year={2018}
}

@inproceedings{mangalam2020not,
  title={\href{https://www.ecva.net/papers/eccv_2020/papers_ECCV/papers/123470749.pdf}{It is not the journey but the destination: Endpoint conditioned trajectory prediction}},
  author={Mangalam, Karttikeya and Girase, Harshayu and Agarwal, Shreyas and Lee, Kuan-Hui and Adeli, Ehsan and Malik, Jitendra and Gaidon, Adrien},
  booktitle={European Conference on Computer Vision},
  pages={759--776},
  year={2020},
  organization={Springer}
}

@inproceedings{sadeghian2019sophie,
  title={\href{https://openaccess.thecvf.com/content_CVPR_2019/papers/Sadeghian_SoPhie_An_Attentive_GAN_for_Predicting_Paths_Compliant_to_Social_CVPR_2019_paper.pdf}{Sophie: An attentive gan for predicting paths compliant to social and physical constraints}},
  author={Sadeghian, Amir and Kosaraju, Vineet and Sadeghian, Ali and Hirose, Noriaki and Rezatofighi, Hamid and Savarese, Silvio},
  booktitle={Conference on Computer Vision and Pattern Recognition},
  pages={1349--1358},
  year={2019}
}

@article{8765346,
  author = {Z. {Cao} and G. {Hidalgo Martinez} and T. {Simon} and S. {Wei} and Y. A. {Sheikh}},
  journal = {IEEE Transactions on Pattern Analysis and Machine Intelligence},
  title = {\href{https://www.computer.org/csdl/journal/tp/2021/01/08765346/1bJTv2i5XJS}{OpenPose: Realtime Multi-Person 2D Pose Estimation using Part Affinity Fields}},
  year = {2019}
}

@article{fang2022alphapose,
  title={\href{https://arxiv.org/pdf/2211.03375.pdf}{Alphapose: Whole-body regional multi-person pose estimation and tracking in real-time}},
  author={Fang, Hao-Shu and Li, Jiefeng and Tang, Hongyang and Xu, Chao and Zhu, Haoyi and Xiu, Yuliang and Li, Yong-Lu and Lu, Cewu},
  journal={IEEE Transactions on Pattern Analysis and Machine Intelligence},
  year={2022},
  publisher={IEEE}
}

@inproceedings{maji2022yolo,
  title={\href{https://arxiv.org/pdf/2204.06806.pdf}{YOLO-Pose: Enhancing YOLO for Multi Person Pose Estimation Using Object Keypoint Similarity Loss}},
  author={Maji, Debapriya and Nagori, Soyeb and Mathew, Manu and Poddar, Deepak},
  booktitle={Conference on Computer Vision and Pattern Recognition},
  pages={2637--2646},
  year={2022}
}

@misc{movenet, title={\href{https://www.tensorflow.org/hub/tutorials/movenet}{MoveNet: Ultra fast and accurate pose detection model.}}, url={https://www.tensorflow.org/hub/tutorials/movenet}, journal={TensorFlow}, publisher={Google}, year={2022}, author={tensorflow.org}}

@misc{posenet, title={\href{https://blog.tensorflow.org/2018/05/real-time-human-pose-estimation-in.html}{Real-time human pose estimation in the browser with tensorflow.js}}, url={https://blog.tensorflow.org/2018/05/real-time-human-pose-estimation-in.html}, journal={The TensorFlow Blog}, publisher={Google}, year={2018}, author={tensorflow.org}}

@inproceedings{yuan2020dlow,
  title={\href{https://www.ecva.net/papers/eccv_2020/papers_ECCV/papers/123540324.pdf}{Dlow: Diversifying latent flows for diverse human motion prediction}},
  author={Yuan, Ye and Kitani, Kris},
  booktitle={European Conference on Computer Vision},
  pages={346--364},
  year={2020},
  organization={Springer}
}

@inproceedings{zhang2021we,
  title={\href{https://openaccess.thecvf.com/content/CVPR2021/papers/Zhang_We_Are_More_Than_Our_Joints_Predicting_How_3D_Bodies_CVPR_2021_paper.pdf}{We are more than our joints: Predicting how 3d bodies move}},
  author={Zhang, Yan and Black, Michael J and Tang, Siyu},
  booktitle={Conference on Computer Vision and Pattern Recognition},
  pages={3372--3382},
  year={2021}
}

@inproceedings{mao2020history,
  title={\href{https://www.ecva.net/papers/eccv_2020/papers_ECCV/papers/123590460.pdf}{History repeats itself: Human motion prediction via motion attention}},
  author={Mao, Wei and Liu, Miaomiao and Salzmann, Mathieu},
  booktitle={European Conference on Computer Vision},
  pages={474--489},
  year={2020},
  organization={Springer}
}

@inproceedings{carion2020end,
  title={\href{https://www.ecva.net/papers/eccv_2020/papers_ECCV/papers/123460205.pdf}{End-to-end object detection with transformers}},
  author={Carion, Nicolas and Massa, Francisco and Synnaeve, Gabriel and Usunier, Nicolas and Kirillov, Alexander and Zagoruyko, Sergey},
  booktitle={European Conference on Computer Vision},
  pages={213--229},
  year={2020},
  organization={Springer}
}

@inproceedings{sun2021rsn,
  title={\href{https://www.computer.org/csdl/proceedings-article/cvpr/2021/450900f721/1yeLfjmzCy4}{Rsn: Range sparse net for efficient, accurate lidar 3d object detection}},
  author={Sun, Pei and Wang, Weiyue and Chai, Yuning and Elsayed, Gamaleldin and Bewley, Alex and Zhang, Xiao and Sminchisescu, Cristian and Anguelov, Dragomir},
  booktitle={Conference on Computer Vision and Pattern Recognition},
  pages={5725--5734},
  year={2021}
}

@inproceedings{ivanovic2018generative,
  title={\href{https://ieeexplore.ieee.org/stamp/stamp.jsp?arnumber=8594393}{Generative modeling of multimodal multi-human behavior}},
  author={Ivanovic, Boris and Schmerling, Edward and Leung, Karen and Pavone, Marco},
  booktitle={International Conference on Intelligent Robots and Systems},
  pages={3088--3095},
  year={2018},
  organization={IEEE}
}

@article{ivanovic2020multimodal,
  title={\href{https://arxiv.org/pdf/2008.03880.pdf}{Multimodal deep generative models for trajectory prediction: A conditional variational autoencoder approach}},
  author={Ivanovic, Boris and Leung, Karen and Schmerling, Edward and Pavone, Marco},
  journal={Robotics and Automation Letters},
  volume={6},
  number={2},
  pages={295--302},
  year={2020},
  publisher={IEEE}
}

@inproceedings{gupta2018social,
  title={\href{https://openaccess.thecvf.com/content_cvpr_2018/papers/Gupta_Social_GAN_Socially_CVPR_2018_paper.pdf}{Social gan: Socially acceptable trajectories with generative adversarial networks}},
  author={Gupta, Agrim and Johnson, Justin and Fei-Fei, Li and Savarese, Silvio and Alahi, Alexandre},
  booktitle={Conference on Computer Vision and Pattern Recognition},
  pages={2255--2264},
  year={2018}
}

@article{vaswani2017attention,
  title={\href{https://proceedings.neurips.cc/paper/2017/file/3f5ee243547dee91fbd053c1c4a845aa-Paper.pdf}{Attention is all you need}},
  author={Vaswani, Ashish and Shazeer, Noam and Parmar, Niki and Uszkoreit, Jakob and Jones, Llion and Gomez, Aidan N and Kaiser, {\L}ukasz and Polosukhin, Illia},
  journal={Advances in Neural Information Processing Systems},
  volume={30},
  year={2017}
}

@inproceedings{xu2020ghum,
  title={\href{https://openaccess.thecvf.com/content_CVPR_2020/papers/Xu_GHUM__GHUML_Generative_3D_Human_Shape_and_Articulated_Pose_CVPR_2020_paper.pdf}{Ghum \& ghuml: Generative 3d human shape and articulated pose models}},
  author={Xu, Hongyi and Bazavan, Eduard Gabriel and Zanfir, Andrei and Freeman, William T and Sukthankar, Rahul and Sminchisescu, Cristian},
  booktitle={Conference on Computer Vision and Pattern Recognition},
  pages={6184--6193},
  year={2020}
}

@misc{mp_holistic, title={\href{https://ai.googleblog.com/2020/12/mediapipe-holistic-simultaneous-face.html}{MediaPipe Holistic — Simultaneous Face, Hand and Pose Prediction, on Device}}, url={hhttps://ai.googleblog.com/2020/12/mediapipe-holistic-simultaneous-face.html}, journal={Google AI Blog}, publisher={Google}, year={2018}, author={Grishchenko, Ivan and Bazarevsky, Valentin}}

@inproceedings{lin2014microsoft,
  title={Microsoft coco: Common objects in context},
  author={Lin, Tsung-Yi and Maire, Michael and Belongie, Serge and Hays, James and Perona, Pietro and Ramanan, Deva and Doll{\'a}r, Piotr and Zitnick, C Lawrence},
  booktitle={Computer Vision--ECCV 2014: 13th European Conference, Zurich, Switzerland, September 6-12, 2014, Proceedings, Part V 13},
  pages={740--755},
  year={2014},
  organization={Springer}
}

@inproceedings{lee2019set,
  title={Set transformer: A framework for attention-based permutation-invariant neural networks},
  author={Lee, Juho and Lee, Yoonho and Kim, Jungtaek and Kosiorek, Adam and Choi, Seungjin and Teh, Yee Whye},
  booktitle={International conference on machine learning},
  pages={3744--3753},
  year={2019},
  organization={PMLR}
}
}

\end{document}